\documentclass[conference]{IEEEtran}
\IEEEoverridecommandlockouts

\usepackage{cite}
\usepackage{amsmath,amssymb,amsfonts}
\usepackage{graphicx}
\usepackage{textcomp}
\usepackage{xcolor}
\def\BibTeX{{\rm B\kern-.05em{\sc i\kern-.025em b}\kern-.08em
    T\kern-.1667em\lower.7ex\hbox{E}\kern-.125emX}}

\usepackage{tikz}
\usepackage{pgfplots}
\usepackage{subcaption}
\usepackage{algorithm}
\usepackage{algpseudocode}

\pgfplotsset{compat=1.11,
    /pgfplots/ybar legend/.style={
    /pgfplots/legend image code/.code={%
       \draw[##1,/tikz/.cd,yshift=-0.25em]
        (0cm,0cm) rectangle (3pt,0.8em);},
   },
}

\pgfplotsset{
    /pgfplots/layers/Bowpark/.define layer set={
        axis background,axis grid,main,axis ticks,axis lines,axis tick labels,
        axis descriptions,axis foreground
    }{/pgfplots/layers/standard},
}

\definecolor{pa1}{RGB}{27,161,226}
\definecolor{pa2}{RGB}{15,153,51}
\definecolor{pa3}{RGB}{216,0,115}
\definecolor{pa4}{RGB}{240,150,9}
\definecolor{pa5}{RGB}{162,0,255}
\definecolor{pa6}{RGB}{0,171,169}
\definecolor{pa7}{RGB}{0,200,200}

\begin{document}

\newcommand{\VSMT}{NB-SMT}
\newcommand{\ARCH}{SySMT}

\title{Non-Blocking Simultaneous Multithreading:\\Embracing the Resiliency of Deep Neural Networks}

\author{\IEEEauthorblockN{Gil Shomron}
\IEEEauthorblockA{\textit{Faculty of Electrical Engineering} \\
\textit{Technion --- Israel Institute of Technology}\\
gilsho@campus.technion.ac.il}
\and
\IEEEauthorblockN{Uri Weiser}
\IEEEauthorblockA{\textit{Faculty of Electrical Engineering} \\
\textit{Technion --- Israel Institute of Technology}\\
uri.weiser@ee.technion.ac.il}
}

\maketitle

\begin{abstract}
Deep neural networks (DNNs) are known for their inability to utilize underlying hardware resources due to hardware susceptibility to sparse activations and weights.
Even in finer granularities, many of the non-zero values hold a portion of zero-valued bits that may cause inefficiencies when executed on hardware.
Inspired by conventional CPU simultaneous multithreading (SMT) that increases computer resource utilization by sharing them across several threads, we propose non-blocking SMT (\VSMT{}) designated for DNN accelerators.
Like conventional SMT, \VSMT{} shares hardware resources among several execution flows.
Yet, unlike SMT, \VSMT{} is non-blocking, as it handles structural hazards by exploiting the algorithmic resiliency of DNNs.
Instead of opportunistically dispatching instructions while they wait in a reservation station for available hardware, \VSMT{} temporarily reduces the computation precision to accommodate all threads at once, enabling a non-blocking operation.
We demonstrate \VSMT{} applicability using \ARCH{}, an \VSMT{}-enabled output-stationary systolic array (OS-SA).  
Compared with a conventional OS-SA, a 2-threaded \ARCH{} consumes 1.4$\times$ the area and delivers 2$\times$ speedup with 33\% energy savings and less than 1\% accuracy degradation of state-of-the-art CNNs with ImageNet.
A 4-threaded \ARCH{} consumes 2.5$\times$ the area and delivers, for example, 3.4$\times$ speedup and 39\% energy savings with 1\% accuracy degradation of 40\%-pruned ResNet-18.

\end{abstract}


\section{Introduction}
Deep neural networks (DNNs) are built of layers that primarily perform dot product operations between activations and weights.
These basic operations are at the core of DNNs that achieve state-of-the-art results in different domains \cite{hannun2014deep, girshick2014rich, silver2016mastering}.
Yet, DNNs comprise abundant computations; for example, state-of-the-art convolutional neural networks (CNNs) may require billions of multiply-and-accumulate (MAC) operations to classify a single image \cite{hannun2014deep, sze2017efficient}.
Their great potential and computational burden have been a fertile ground for research and development of efficient DNN hardware accelerators over the last decade \cite{reuther2019survey, jouppi2017datacenter, chen2014dadiannao}.

The control flow of DNNs is mostly predictable, yet computations are still executed inefficiently on underlying hardware.
For example, DNNs may consist of many zero-valued activations and weights \cite{albericio2016cnvlutin}.
During inference, a layer output is usually followed by a ReLU activation function, which clamps negative activation values to zero \cite{nair2010rectified}.
In addition, static pruning techniques push the limits of model sparsity by zeroing out insignificant weights \cite{han2015learning, li2016pruning}.
Zeros can be also found in finer granularities \cite{lascorz2019shapeshifter}; a quantized 8-bit DNN has many values that can be effectively represented only by the 4-bit least-significant bits (LSBs).
This unstructured sparsity can be leveraged to increase efficiency, thereby improving performance and reducing energy.
Until now, DNN accelerators have handled such inefficiencies with compressed encodings \cite{gondimalla2019sparten, parashar2017scnn, han2016eie}, output zero-value prediction \cite{akhlaghi2018snapea, song2018prediction, shomron2019thanks}, input zero-value skipping \cite{sharify2019laconic, kim2017zena, zhang2016cambricon, albericio2016cnvlutin}, and working with bit-serial schemes \cite{sharify2019laconic, judd2016stripes}.



In this paper, we introduce \emph{non-blocking simultaneous multithreading} (\VSMT{}), a new approach to tackle sparsity and increase hardware efficiency.
Conceptually, \VSMT{} is based on the well-known SMT used to concurrently execute multiple instruction flows on shared resources \cite{yamamoto1994performance, yamamoto1995increasing, tullsen1995simultaneous, eggers1997simultaneous}.
In the same manner that SMT keeps several hardware threads to increase utilization of hardware resources, we propose maintaining a number of ``DNN threads'' that run in parallel so as to increase utilization of DNN hardware resources.

Conventional SMT dispatches instructions to an execution unit in an opportunistic manner.
That is, if instruction dependencies are met and its needed resources are available, it will be executed; otherwise, the instruction will wait in a reservation station.
The \VSMT{} scheme employed in this paper avoids this online scheduling by ``squeezing'' two (or more) threads together to the shared resource (e.g., execution unit) by temporarily reducing their numerical precision.
By doing so, we (1) leverage DNN tolerance to reduced numerical precision, thereby enabling a non-blocking operation; 
(2) do not break the systematic operation of DNNs, thereby enabling implementation of SMT in dataflow architectures, which are popular as DNN accelerators;
and (3) achieve a speedup that is directly proportional to the number of threads.



\VSMT{} may be implemented in different DNN accelerator architectures and can support concurrent execution of threads that originate from different models or from within the same model.
In this paper, we focus on the latter and demonstrate 2-threaded and 4-threaded \VSMT{} as an extension to an 8-bit output-stationary (OS) systolic array (SA) for matrix multiplication \cite{kung1979systolic, shomron2019smt, gupta2015deep}, which we named \ARCH{}.  
Compared with the conventional OS-SA, a 2-threaded \ARCH{} achieves a 2$\times$ speedup with 33\% energy reduction and less than 1\% accuracy degradation of state-of-the-art CNNs with a 1.4$\times$ area increase.
As for 4-threads, we observe that some layers contribute more errors to inference than others when executed with \VSMT{}.
Therefore, we trade speedup for accuracy by decreasing the number of running threads in selective layers.
Given a 1\% accuracy degradation cap, a 4-threaded \ARCH{} delivers, for example, 3.4$\times$ speedup with 37\% energy reduction and 2.5$\times$ area increase with 40\%-pruned ResNet-18, compared with the conventional OS-SA.

Our contributions in this paper are as follows:
\begin{itemize}
  \item We introduce the concept of non-blocking simultaneous multithreading (\VSMT{}), which increases DNN hardware utilization by exploiting DNN algorithmic resiliency and unstructured sparsity.
  Specifically, we present an \VSMT{} scheme in which the non-blocking operation is enabled by reducing the numerical precision of values on-the-fly.
  By not blocking any thread, \VSMT{} achieves a speedup that is directly proportional to the number of threads.
  
       

  \item We demonstrate \VSMT{} applicability using \ARCH{}, which is an \VSMT{}-enabled output-stationary systolic array.
        We describe different resource sharing strategies in which \ARCH{} employs both MAC unit and output register sharing.
  \item We evaluate a 2-threaded and a 4-threaded \ARCH{} in terms of speedup, area, power, energy, and model accuracy with various state-of-the-art CNN models and the ImageNet dataset.
\end{itemize}

The rest of this paper is organized as follows:
Section~\ref{sec:motivation} describes the rationale behind \VSMT{},
Section~\ref{sec:the_idea} presents the basic principals of \VSMT{},
Section~\ref{sec:sa} demonstrates \VSMT{} as an extension to an output-stationary systolic array (\ARCH{}),
Section~\ref{sec:eval} evaluates the impact of \VSMT{} on \ARCH{} implementation as well as on model accuracy,
Section~\ref{sec:related_work} discusses the applicability of \VSMT{} in other accelerators and reviews related work,
and Section~\ref{sec:conclusions} concludes.

\section{Motivation}
\label{sec:motivation}

The CPU instruction pipeline faces many challenges in achieving efficient execution.
These inefficiencies, also known as hazards, originate from the application's dynamic execution flow and from the generality of the architecture (i.e., general-purpose).
DNNs, on the other hand, work in a systematic, layer-by-layer fashion, with mostly MAC operations taking place during inference, making their control and data flow deterministic; which and how many computations will be conducted, what is the model's memory footprint, where are weights stored, and where will activations be stored during execution, can all be deduced prior to execution (neglecting special cases of conditional DNNs, for example).
Yet, DNNs still exhibit inefficiencies when considering the actual values that propagate through the layers.

\textbf{Sparsity.}
DNNs comprise zero-valued activations and weights \cite{nikolic2019characterizing}.
Zero-valued activations are produced dynamically during inference, due, among other things, to the popular use of the ReLU activation function, which clamps negative values to zero \cite{nair2010rectified}.
On the other hand, weights are static during inference, and in most cases, not many of them are zero-valued when trained only with a loss function.
However, training the network with L1 regularization or pruning the network, for example, can substantially reduce the number of parameters (i.e., increase the number of zero-valued weights) with negligible decrease in model accuracy \cite{li2016pruning}.
For example, 60\% of ResNet-50 parameters can be discarded \cite{liu2018rethinking} by iteratively trimming small weights and retraining the model in an unstructured manner \cite{han2015learning}.

\textbf{Partial sparsity.}
Zeros can be also observed when looking within the numerical representation.   
DNN tensors usually follow a bell-shaped distribution, such as Gaussian or Laplace \cite{banner2019post}.
Therefore, when considering a quantized DNN, some values will only be represented by a portion of the LSBs, leaving the most-significant bits (MSBs) to be equal to zero \cite{lascorz2019shapeshifter}.
Throughout this paper we use 8-bit model representations, so by ``partial sparsity'' we refer to those numbers that can be represented solely by 4 bits.


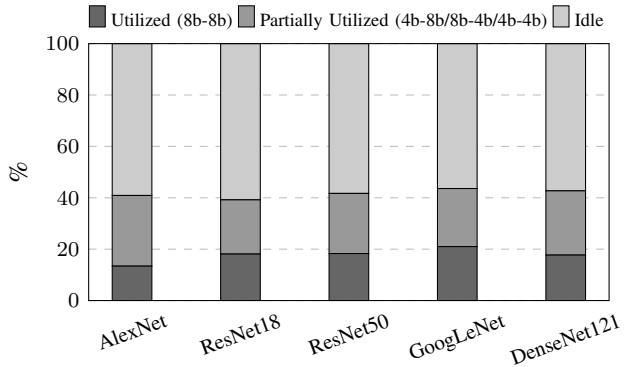
\begin{figure}[t] 
	\centering

	\begin{tikzpicture}
	\begin{axis}[
			ybar stacked, bar width=15pt,
			width=8.5cm, height=5cm,
	    	enlarge x limits=0.1, enlarge y limits=0,
	    	xlabel near ticks, ylabel near ticks,
	    	legend style={at={(0.5,-0.20)}, anchor=north,legend columns=-1},
	    	ylabel={\%}, ymin={0}, ymax={100},
	    	symbolic x coords={AlexNet, ResNet18, ResNet50, GoogLeNet, DenseNet121},
	    	xtick=data,
	    	x tick label style = {font=\footnotesize, rotate=20},
			y tick label style = {font=\footnotesize},
			legend style={font=\scriptsize, legend columns=4, legend cell align=left, at={(0.5, 1.01)}, anchor=south, draw=none},
			tick style={draw=none},
			xmajorgrids, ymajorgrids, major grid style={dashed},
			set layers=Bowpark
	    ]
	    
		\addplot[fill=black!60] plot coordinates {(AlexNet,13.4) (ResNet18,18.1) (ResNet50,18.3) (GoogLeNet,21.0) (DenseNet121,17.7)};
		\addplot[fill=black!40] plot coordinates {(AlexNet,27.5) (ResNet18,21.1) (ResNet50,23.4) (GoogLeNet,22.6) (DenseNet121,25.0)};
		\addplot[fill=black!20] plot coordinates {(AlexNet,59.1) (ResNet18,60.8) (ResNet50,58.3) (GoogLeNet,56.4) (DenseNet121,57.3)};

		\legend{Utilized (8b-8b), Partially Utilized (4b-8b/8b-4b/4b-4b), Idle}
		
	\end{axis}
	\end{tikzpicture}

\caption{Utilization of 8-bit MAC units during CNN inference, as simulated given the entire ILSVRC-2012 dataset \cite{russakovsky2015imagenet}.
         On average, only 20\% of the MAC units are fully utilized.}
\label{fig:mac_util}
\end{figure}

\textbf{Unstructured sparsity.}
Activation sparsity is unstructured by nature, as the zero-valued activations may be scattered without any confined structure.
Moreover, the values themselves are input-dependent, and thereby dynamic.
Weights, on the other hand, are static during inference and therefore can be pruned either in an unstructured or structured manner.
A general rule of thumb is that unstructured pruning techniques achieve a better parameter reduction to accuracy reduction ratio than do structured techniques.
Indeed, with unstructured pruning, the algorithm has the freedom to cancel parameters in weight granularity, whereas structured pruning algorithms are constrained to remove parameters in larger granularity, such as channels or filters \cite{wen2016learning}.
The downside of unstructured pruning is, however, that it is not easily exploited by hardware \cite{liu2018rethinking}.

The unstructured sparse inputs cause spontaneous underutilization of the MAC units.
From a hardware perspective, a MAC unit with one of its inputs equals to zero is practically idle; and an 8b-8b MAC unit with an effective input data-width of 4 bits is only partially utilized.
Figure~\ref{fig:mac_util} presents the average MAC utilization of five popular CNN models.
We observe that, on average, 60\% of MAC operations result in idle MAC units, since one of their inputs is zero-valued;
20\% of MAC operations partially utilize the MAC units, since one of their inputs, or both, are effectively represented with 4 bits;
and in a mere 10\% of the time, the MAC operations fully utilize the MAC units.
To increase hardware utilization, we propose non-blocking simultaneous multithreading (\VSMT{}) that exploits both the unstructured sparsities of the activations and weights, as well as DNN tolerance to numerical precision reduction.

\textbf{Algorithmic resiliency.}
DNNs are fault-tolerant \cite{reagen2018ares, li2017understanding}; they can absorb connection removals \cite{han2015learning, shomron2019thanks} and numerical precision reduction \cite{kravchik2019low, shkolnik2020robust} with only a ``graceful degradation'' in performance \cite{kerlirzin1993robustness}.
For example, DNNs can be quantized from FP32 to INT8 in a straight-forward post-training min-max uniform quantization with no significant loss in accuracy \cite{krishnamoorthi2018quantizing}.
DNN tolerance can be harnessed in order to ease design constraints.
Specifically, \VSMT{} builds upon DNN resiliency to handle structural hazards without stalling any thread, as opposed to conventional SMT.
Avoiding stalls coalesces with the way DNNs operate during inference.

\textbf{Systematic operation.}
Inference with DNNs is a compute-intensive task that requires minor control.
For example, ResNet-50 \cite{he2016deep} requires 4 billion MAC operations to classify a single 224$\times$224 colored image from the ImageNet dataset \cite{russakovsky2015imagenet}.
During these 4 billion computations, there is not a single control branch --- the entire control flow is predictable.
These application characteristics have driven computer architects to design highly parallel DNN architectures with almost no control logic \cite{jouppi2017datacenter, chen2014dadiannao}.
The lack of control capabilities, which is a consequence of the systematic operation of DNNs, stands in contrast to the conventional SMT way of operation, which may stall threads as a function of the current state of microarchitecture.
By completely avoiding stalls, we enable an SMT implementation in DNN hardware architectures.

%

\section{\VSMT{}: The Basic Idea}
\label{sec:the_idea}

Conventional SMT is based on the observation that a single thread might not fully utilize the execution resources.
SMT tries to increase utilization by exploiting thread-level parallelism, that is, dispatching instructions from different threads to the same resources.
Inspired by SMT, we propose \VSMT{}, a special ``SMT'' designated for the environment of DNNs.

\VSMT{} is conceptually similar to traditional SMT, in the sense that the context of more than one thread is kept on hardware in parallel.
In all other aspects, however, \VSMT{} differs from traditional SMT:
first, it compensates for underutilization caused by particular data \emph{values};   
and second, it is non-blocking.
The exact \VSMT{} implementation may vary, depending on the underlying architecture and target DNNs.
In this paper, since we target quantized neural networks, instead of keeping operations waiting in reservation stations on structural hazards, \VSMT{} ``squeezes'' operations to the same hardware by momentarily reducing their precision, considering DNN tolerance to reduction in numerical precision.


\subsection{Hiding Inefficiencies}
MAC unit operation is value-dependent.
For example, let $(X,W)$ be an input pair that consists of two vectors of length $K$ that are to be multiplied.
The process of achieving the result includes $K$ MAC operations of the corresponding elements of $X$ and $W$, that is, $O = \sum_{i=0}^{K-1} x_i w_i$.
Now, assume $X$ comprises 50\% zero-valued elements.
In this case, 50\% of MAC operations are effectively redundant, as $0\times x = 0$ and $0 + x = x$.

\VSMT{} increases utilization with additional threads that exploit the idle MAC units.
For example, a 2-threaded (2T) \VSMT{} will include two independent input pairs $(X,W)_1$ and $(X,W)_2$, each of which will produce a result of its own, $O_1$ and $O_2$, respectively.
Thus, if the first pair does not require the MAC unit, there is a chance that the second thread will, thereby utilizing it.
To support \VSMT{}, the hardware should include additional data path and registers.
The hardware should also be capable of handling thread collisions, i.e., cases in which the threads' computation demands are higher than the MAC unit capabilities.

\subsection{Thread Collisions}
Thread collisions can be handled with queues and backpressure to support congestions \cite{shomron2019smt}.
However, \VSMT{} takes another path, exploiting DNNs' resiliency and temporarily reducing the threads' numerical precision so that execution units are still able to accommodate all thread computations in that same cycle.
Thread collision incurs reduction in precision which contributes some error to the overall computation, for example, the case of a single 8b-8b MAC unit and two threads with 8b-8b and 8b-8b input pairs.
On the other hand, for example, threads that are represented solely by their 4-bit LSBs can share the underlying MAC unit without affecting the original computation.
We demonstrate these scenarios next.

\subsection{Squeezing Them In}

\begin{figure}[t]

	\begin{subfigure}[b]{\columnwidth}
	  \centering
	  \includegraphics[width=\linewidth]{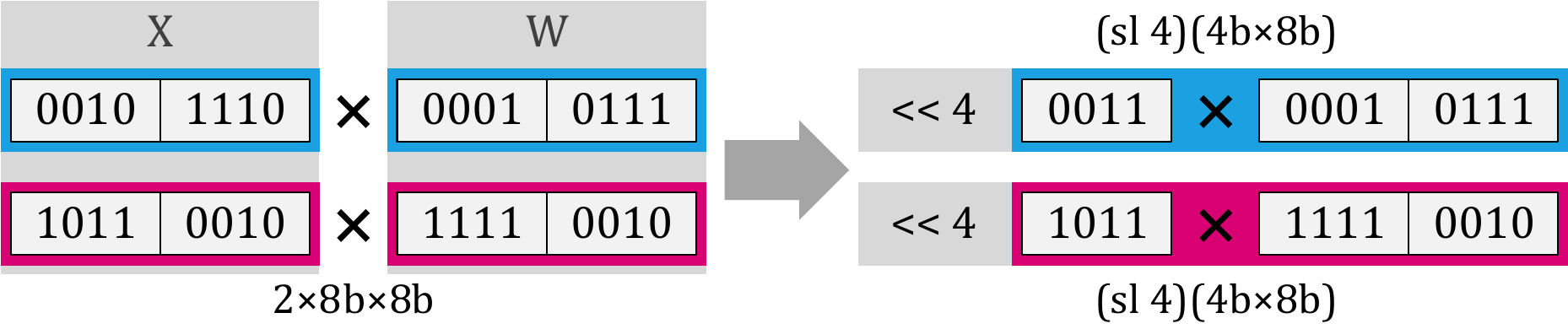}  
	  \caption{\textbf{Precision reduction:} Both input pairs require the multiplier.
	           Therefore, without loss of generality, $X$ inputs are approximated by their 4-bit MSBs (rounded first) to achieve two 4b-8b multiplications.}
	  \label{fig:smt_example:1}
	\end{subfigure}
	
	\par\medskip
	
	\begin{subfigure}[b]{\columnwidth}
	  \centering
	  \includegraphics[width=\linewidth]{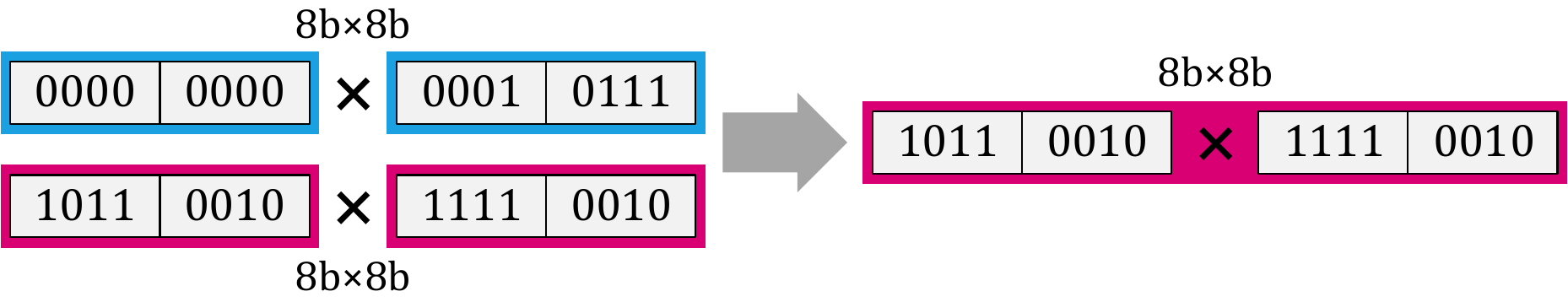}  
	  \caption{\textbf{8-bit sparsity}: First thread has a zero-valued operand.
	                        The second thread can, therefore, use the entire 8b-8b multiplier.}
	  \label{fig:smt_example:2}
	\end{subfigure}
	
	\par\medskip
	
	\begin{subfigure}{\columnwidth}
	  \centering
	  \includegraphics[width=\linewidth]{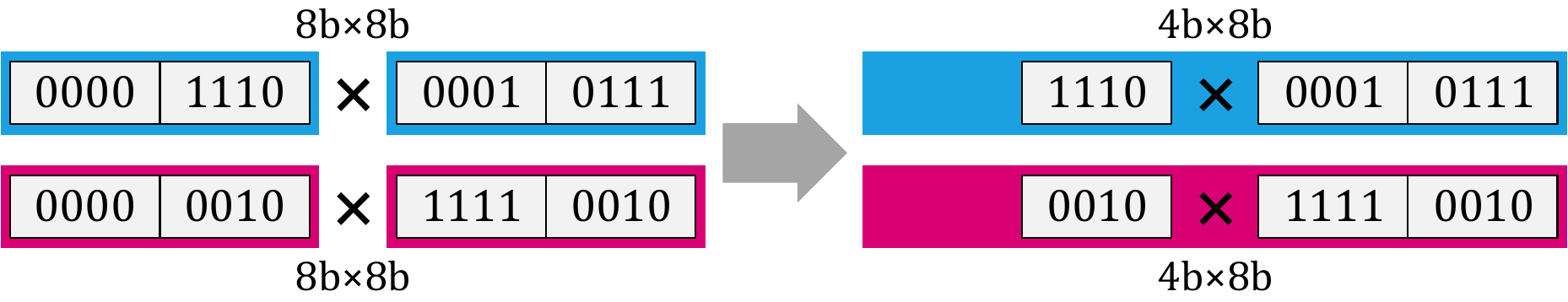}  
	  \caption{\textbf{Activations' 4-bit sparsity:} $X$ inputs may be effectively represented by their 4-bit LSBs.
	           Two 4b-8b multiplications, therefore, take place with no computation error.}
	  \label{fig:smt_example:3}
	\end{subfigure}
	
	\par\medskip
	
	\begin{subfigure}{\columnwidth}
	  \centering
	  \includegraphics[width=\linewidth]{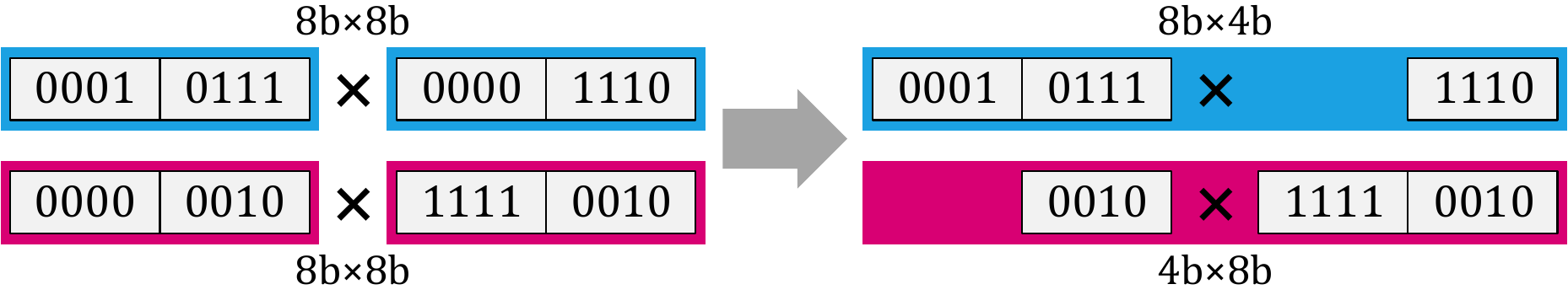}  
	  \caption{\textbf{Activation and weight 4-bit sparsity:} Considering data-width of both inputs to avoid precision reduction.}
	  \label{fig:smt_example:4}
	\end{subfigure}
	
	\par\medskip
	
	\begin{subfigure}{\columnwidth}
	  \centering
	  \includegraphics[width=\linewidth]{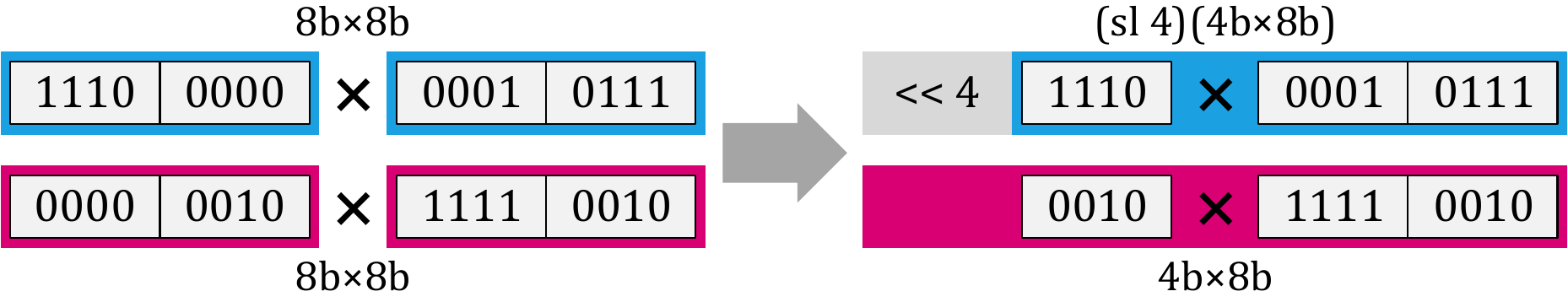}  
	  \caption{\textbf{Activations' 4-bit sparsity with reduced precision:} A combination of (a) and (d).}
	  \label{fig:smt_example:5}
	\end{subfigure}

\caption{Examples of how \VSMT{} ``squeezes'' inputs to a single flexible multiplier unit, capable of either one 8b-8b multiplication or two independent 4b-8b multiplications.}
\label{fig:smt_example}
\end{figure}

Consider a flexible multiplier, capable of conducting either a single 8b-8b multiplication or two 4b-8b multiplications per cycle (we further elaborate on such flexible multipliers in Section~\ref{sec:fmac}).
For simplicity's sake, throughout this section we consider only two threads, that is, two input pairs, $(X,W)_1$ and $(X,W)_2$, with unsigned values.


\subsubsection{Precision Reduction}
\label{sec:the_idea_prec_reduce}
On-the-fly precision reduction truncates values represented by more than 4 bits to 4 bits.
Reducing thread precision takes place when a thread collision occurs and the thread operands are represented by more than 4 bits.
Before reducing the 8-bit value (activation or weight) to 4 bits, we round the number to the nearest integer that is a whole multiple of 16 ($2^4$), to mitigate the noise it contributes to the entire inference pass.


Figure~\ref{fig:smt_example:1} illustrates an example of reducing the numerical precision of the activation values (without loss in generality).
$(X,W)_1$ and $(X,W)_2$ are equal to $(46_{10}, 23_{10})$ and $(178_{10}, 242_{10})$, respectively.
Both $X$ inputs MSBs are rounded and truncated so as to be represented by 4-bits, yielding $(3_{10}, 23_{10})$ and $(11_{10}, 242_{10})$, respectively.
Two 4b-8b multiplications then take place, followed by a 4-bit shift left, resulting in two approximated results $1104_{10}$ (instead of $1058_{10}$) and $42592_{10}$ (instead of $43076_{10}$).
It is obvious, however, that it is unnecessary to reduce precision of all input values.

\subsubsection{8-Bit Sparsity}
If $X_1$ or $W_1$ or both are zero-valued, $(X,W)_2$ can use the entire 8b-8b multiplier, and vice versa.
For example, consider the two input pairs in Fig.~\ref{fig:smt_example:2}, $(0, 23_{10})$ and $(178_{10}, 242_{10})$.
It is clear that the first thread does not require the multiplier, since its first multiplication operand is $0$.
The second thread will, therefore, utilize the entire multiplier to produce the original result with no computation error.

\subsubsection{4-Bit Sparsity}
If both threads are effectively represented by 4b-8b or 4b-4b, computation error is avoided.
Without loss in generality, we consider only the 4-bit representation of $X$ inputs.
Figure~\ref{fig:smt_example:3} illustrates an example of thread collision.
The easiest way to solve the collision is simply by considering only the 4-bit MSBs, as described in Fig.~\ref{fig:smt_example:1}.
Instead, we observe that in both threads, the four MSB bits are zero-valued.
Therefore, instead of trimming the threads' LSBs, we keep them, taking into account that, in this case, multiplication should not be followed by a shift left operation.

4-bit sparsity of both $X$ and $W$ inputs may be exploited as well, as depicted in Fig.~\ref{fig:smt_example:4}.
In this example, the $X$ and $W$ of the first thread are swapped.
Now, the $W$ input of the first thread uses the LSBs, neglecting the zero-valued 4-bit MSBs.
Even though exploiting data-width variability of both inputs seems trivial, additional hardware is required to dynamically determine which of the inputs, $X$ or $W$, will enter the 4-bit multiplier port.

Figure~\ref{fig:smt_example:5} illustrates an example in which 4-bit sparsity and precision reductions are needed.
In this example, the first and second threads effectively use 8b-8b and 4b-8b, respectively.
The precision of the first thread is, therefore, reduced to fit the multiplier.
The values in this example lead to, effectively, no collision, since the 4-bit LSBs of the first thread are all zeros.
If this was not so, error was contributed by the first thread.


\subsection{Shared Resources}

\begin{figure*}[t]
	\begin{subfigure}{0.32\textwidth}
	  \centering
	  \includegraphics[height=2.2cm]{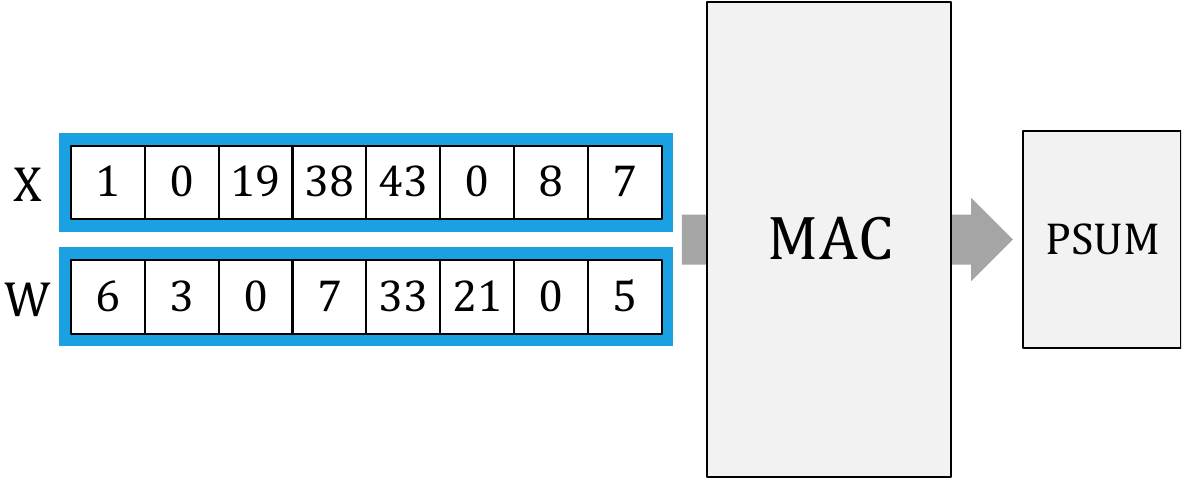}  
	  \caption{Conventional}
	  \label{fig:vector_tiling:1}
	\end{subfigure}
	\begin{subfigure}{0.4\textwidth}
	  \centering
	  \includegraphics[height=2.2cm]{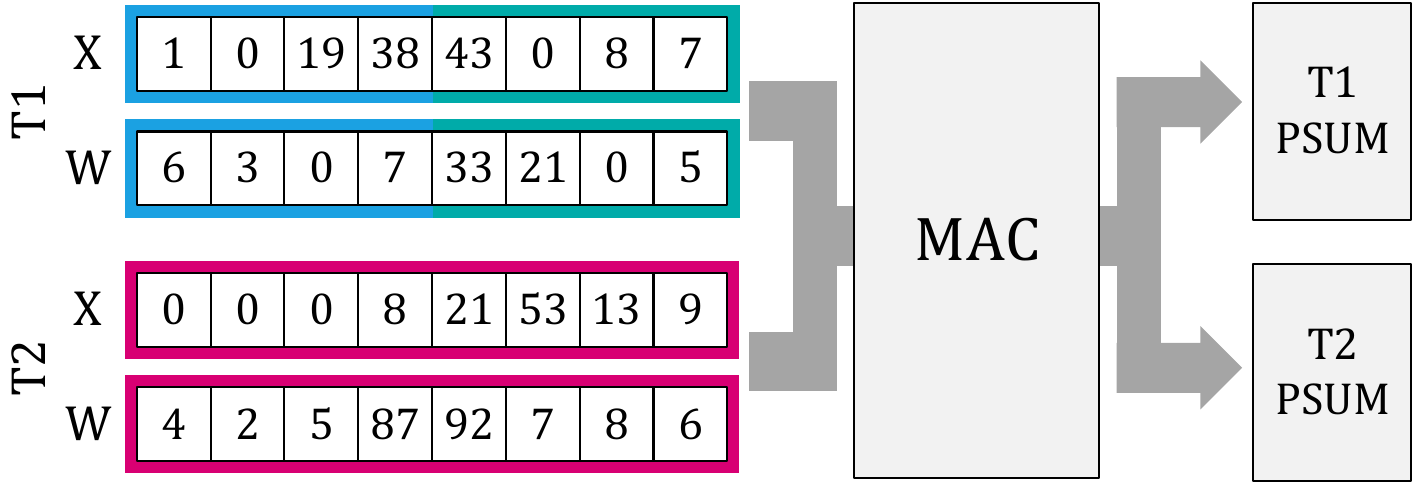}  
	  \caption{Shared MAC, independent psums}
	  \label{fig:vector_tiling:2}
	\end{subfigure}
	\begin{subfigure}{0.28\textwidth}
	  \centering
	  \includegraphics[height=2.2cm]{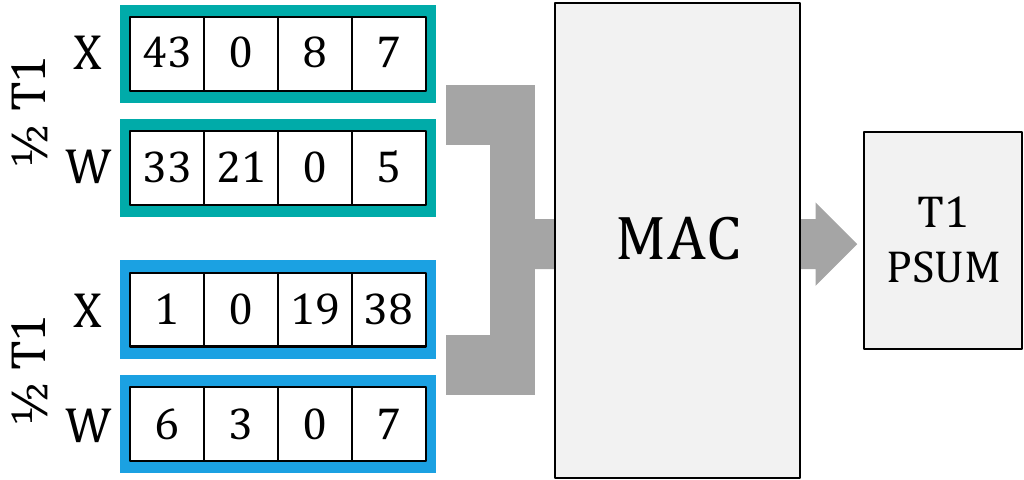}  
	  \caption{Shared MAC, shared psum}
	  \label{fig:vector_tiling:3}
	\end{subfigure}
	
\caption{Shared resources. \textbf{(a) Conventional:} A single input, $(X,W)$, feeds a single MAC unit to produce a single output, $O$, in a single partial sum register (psum).
         \textbf{(b) Shared MAC:} Two independent threads, $(X,W)_1$ and $(X,W)_2$, feed a single MAC unit, producing two independent outputs, $O_1$ and $O_2$, respectively.
         \textbf{(c) Shared MAC and psum:} Two threads, \emph{originating from the same input}, feed a single MAC unit, producing a single output.}
\label{fig:vector_tiling}
\end{figure*}

\VSMT{} can execute several independent threads, i.e., per-thread $X$, $W$, and $O$ (Fig.~\ref{fig:vector_tiling:2}), which is, in a sense, similar to the way conventional SMT operates whereby each hardware thread is executed independently.
Logically, however, threads can be dependent, so instead of independent ``DNN threads'' we propose threads that originate from the same execution flow, somewhat similar to the difference between software threads and processes.
By doing so, we can share not only the MAC unit but also additional resources:
(1) activation registers sharing: the same activation is used with different weights (filters), thereby computing different output activations;
(2) weight registers sharing: the same weight is used with different activations (e.g., batch or another convolution window), thereby computing different output activations;
and (3) output registers sharing: different activations and their corresponding weights compute the same output activation.



We focus here on output sharing.
Let $(X,W)$ be an input pair that consists of two vectors of length $K$ that are to be multiplied.
Obviously, the multiplication result is a scalar equal to $\sum_{i=0}^{K-1} x_i w_i$.
With 2T \VSMT{}, for example, instead of doubling the partial sum registers, we split the input vectors, $X$ and $W$, between the two threads, and so both thread results are summed up to produce a single scalar value.
In other words, given two independent threads, $(X,W)_1$ and $(X,W)_2$, of respective lengths $K_1$ and $K_2$, the outputs are also independent as follows:
\begin{equation}
O_1 = \sum_{i=0}^{K_1-1} x_i^{(1)} w_i^{(1)} \;\;\; \text{and} \;\;\; O_2 = \sum_{i=0}^{K_2-1} x_2^{(2)} w_i^{(2)} \,.
\end{equation}
A single $(X,W)$ input pair may, however, can span two different threads:
\begin{equation}
\begin{aligned}
\text{Thread 1:} \;\;\; &(X[0, K/2-1], W[0, K/2-1]) \\
\text{Thread 2:} \;\;\; &(X[K/2, K-1], W[K/2, K-1]) \,,
\end{aligned}
\label{eq:vec_threads}
\end{equation}
where $G[i_1,i_2]$ denotes an arbitrary vector $G$, such as $X$ or $W$, consisting of elements $i_1$ through $i_2$, and, for simplicity's sake, we assume $K$ is an even number.
Both threads therefore contribute to the same output as follows (see Fig.~\ref{fig:vector_tiling}):
\begin{equation}
O = \sum_{i=0}^{K/2-1} x_i w_i + \sum_{i=K/2}^{K-1} x_i w_i \,.
\label{eq:vec_sum}
\end{equation}

\section{\ARCH{}: Architecture Use Case}
\label{sec:sa}

\VSMT{} may be enabled in different DNN hardware architectures.
Specifically, we use an output-stationary (OS) systolic array (SA) designated for matrix multiplication \cite{kung1979systolic, gupta2015deep} as our case study.
In this section, we first briefly review SAs.
We then describe how data may be organized to decrease the number of thread collisions.
And finally, we present the microarchitecture of \ARCH{} --- an OS-SA \VSMT{} which employs output sharing.
Throughout this paper we focus on the computation core.

\subsection{Output-Stationary Systolic Arrays}
SAs comprise a grid of processing elements (PEs).
PEs work in tandem: each PE independently receives inputs from its upstream PE neighbors, conducts a certain task whose result it stores locally, and forwards its inputs downstream.
The well-defined interactions between adjacent PEs and the specific and confined task that each PE conducts enable efficient data reuse and scalability \cite{kung1979systolic}.

\begin{figure*}[t] 
    \begin{subfigure}[b]{.45\textwidth}
	  \centering
	  \includegraphics[height=4.8cm]{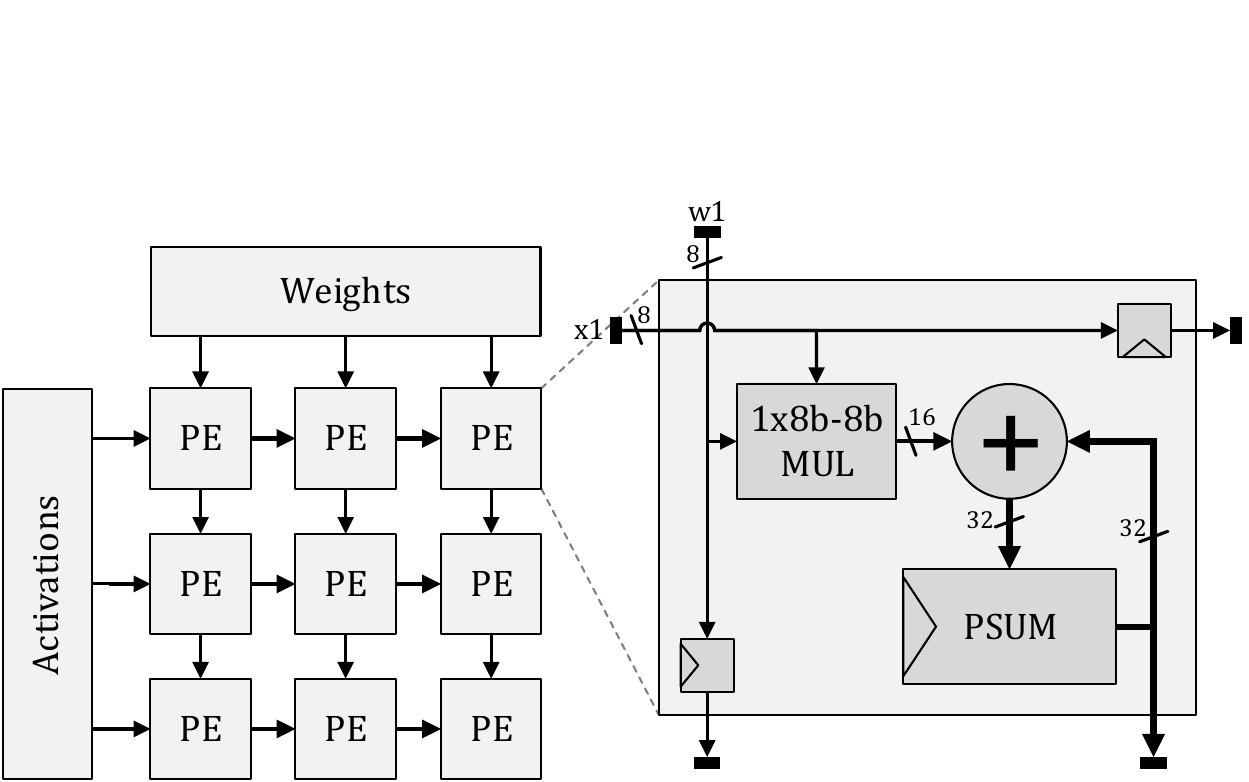}  
	  \caption{Conventional OS-SA}
	  \label{fig:os-sa_arch}
	\end{subfigure}
	\begin{subfigure}[b]{.55\textwidth}
	  \centering
	  \includegraphics[height=4.8cm]{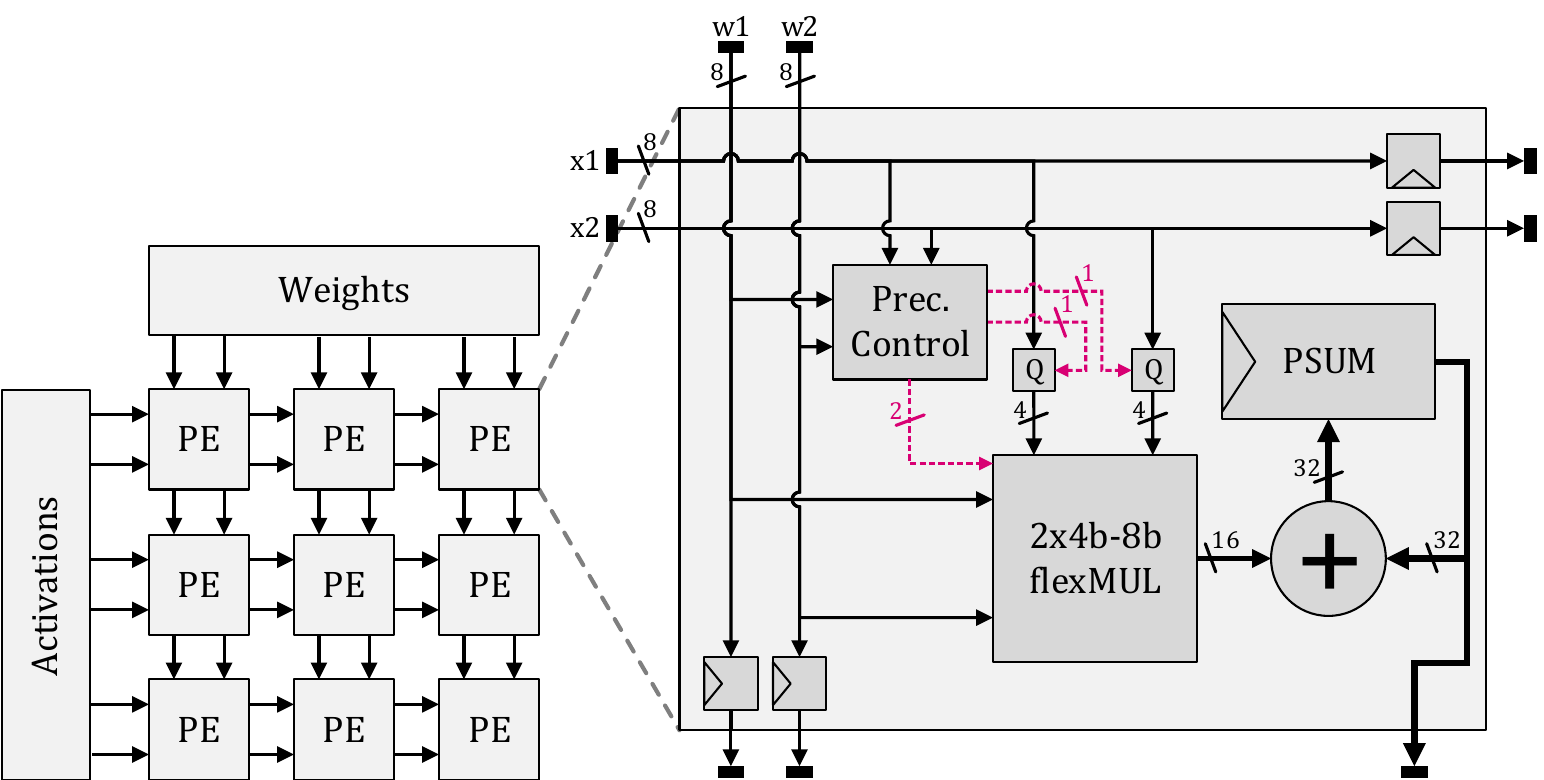}  
	  \caption{2T \ARCH{}}
	  \label{fig:os-sa_mt_arch}
	\end{subfigure}
	\caption{Architecture and microarchitecture of a 3$\times$3 OS-SA and \ARCH{}.}
	\label{fig:arch}
\end{figure*}

SAs serve many applications and come in many shapes and forms.
Specifically, we take an SA designated for matrix multiplication and use it for computation of convolutional layers \cite{chetlur2014cudnn}.
In addition, we use a variant of SA that is OS.
In the OS-SA, each PE receives an activation and weight per cycle and accumulates their multiplication results locally.
Data is pushed to the PE array in a skewed manner, so that corresponding activations and weights meet in the appropriate PE.
Figure~\ref{fig:os-sa_arch} depicts the OS-SA architecture and PE uarch.


The \ARCH{} grid is almost identical to the conventional SA grid, except that connectivity is scaled with the number of threads, as illustrated in Fig.~\ref{fig:os-sa_mt_arch}.
In Fig.~\ref{fig:vector_tiling:3} we illustrate how an input vector is split into two threads; in the same manner, we split the activation and weight input matrices into two threads.
Let $X_{M \times K}$ and $W_{K \times N}$ be the two input matrices, and consider a 2-threaded design, for example.
Each row in $X$ and each column in $W$ is split as described by Eq.~(\ref{eq:vec_threads}).
Each PE is therefore able to perform the computation presented in Eq.~(\ref{eq:vec_sum}).



\subsection{Data Arrangement}
\label{sec:reorder}
Given two threads, it would be ideal if data was arranged so that in each cycle, one thread holds at least one zero-valued term and the other thread does not, or that both threads hold a pair represented by 4b-8b or 4b-4b.
Reordering the data is not, however, trivial:
(1) it is impractical to reorder the activation matrices according to their momentary values, since activation values are dynamic;
(2) weights are static during inference, but we do not expect the weight tensors to exhibit much correlation between rows as we expect from the activation columns, since each row in the input activations matrix represents a sliding window \cite{chetlur2014cudnn}, and activations are known to exhibit spatial correlation \cite{mahmoud2018diffy, shomron2019thanks};
and (3) the SA structure dictates specific scheduling of the data inputs.
Given the two $X_{M \times K}$ and $W_{K \times N}$ input matrices, reordering of $X$ must take place in column granularity followed by reordering of the corresponding $W$ rows accordingly so as to maintain SA data scheduling.

\begin{figure}[t]
\centering
	\begin{subfigure}[b]{\columnwidth}
	  \centering
	  \includegraphics[height=3.7cm]{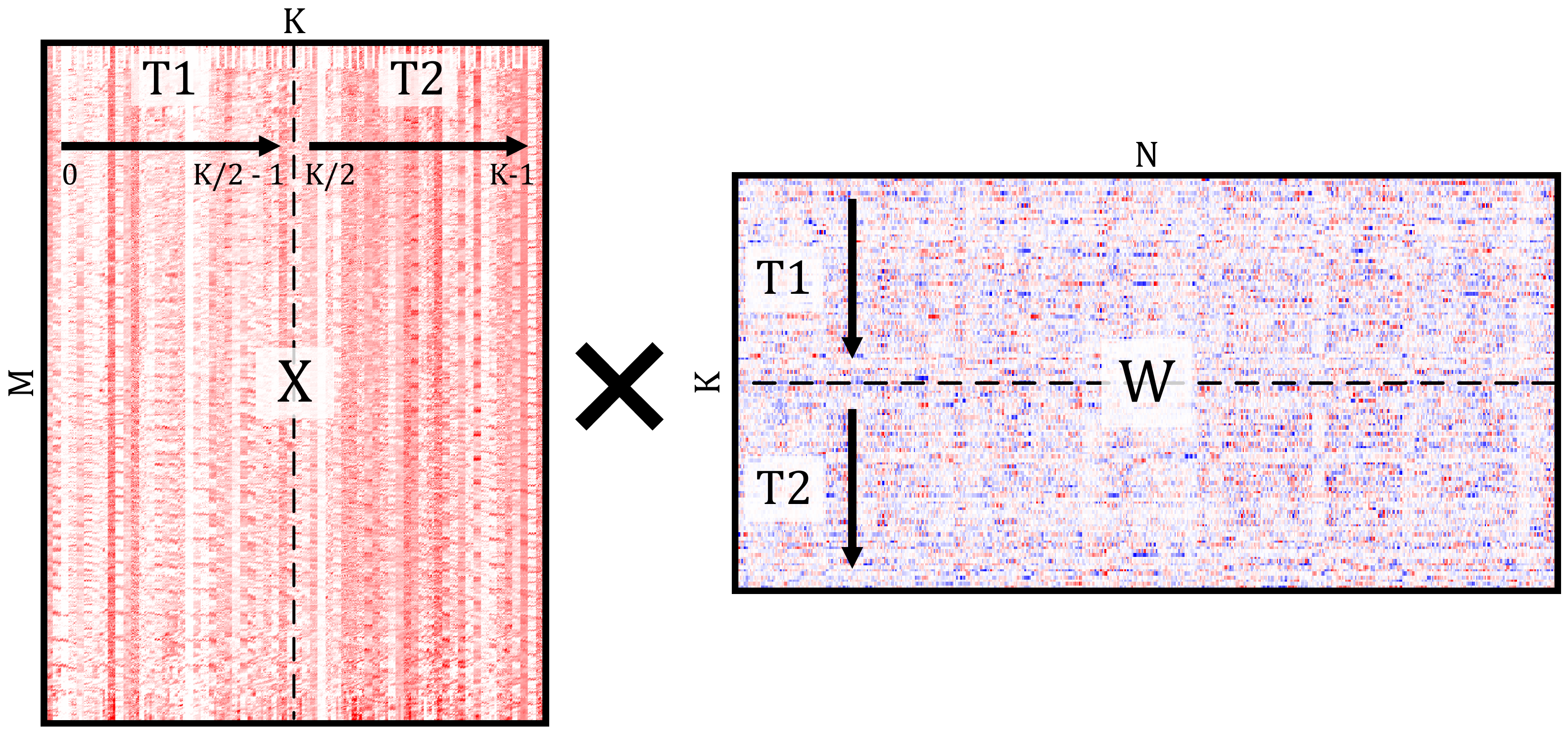}
	  \caption{Before reordering}
	  \label{fig:reorder:before}
	\end{subfigure}

	\begin{subfigure}[b]{\columnwidth}
	  \centering
	  \includegraphics[height=3.7cm]{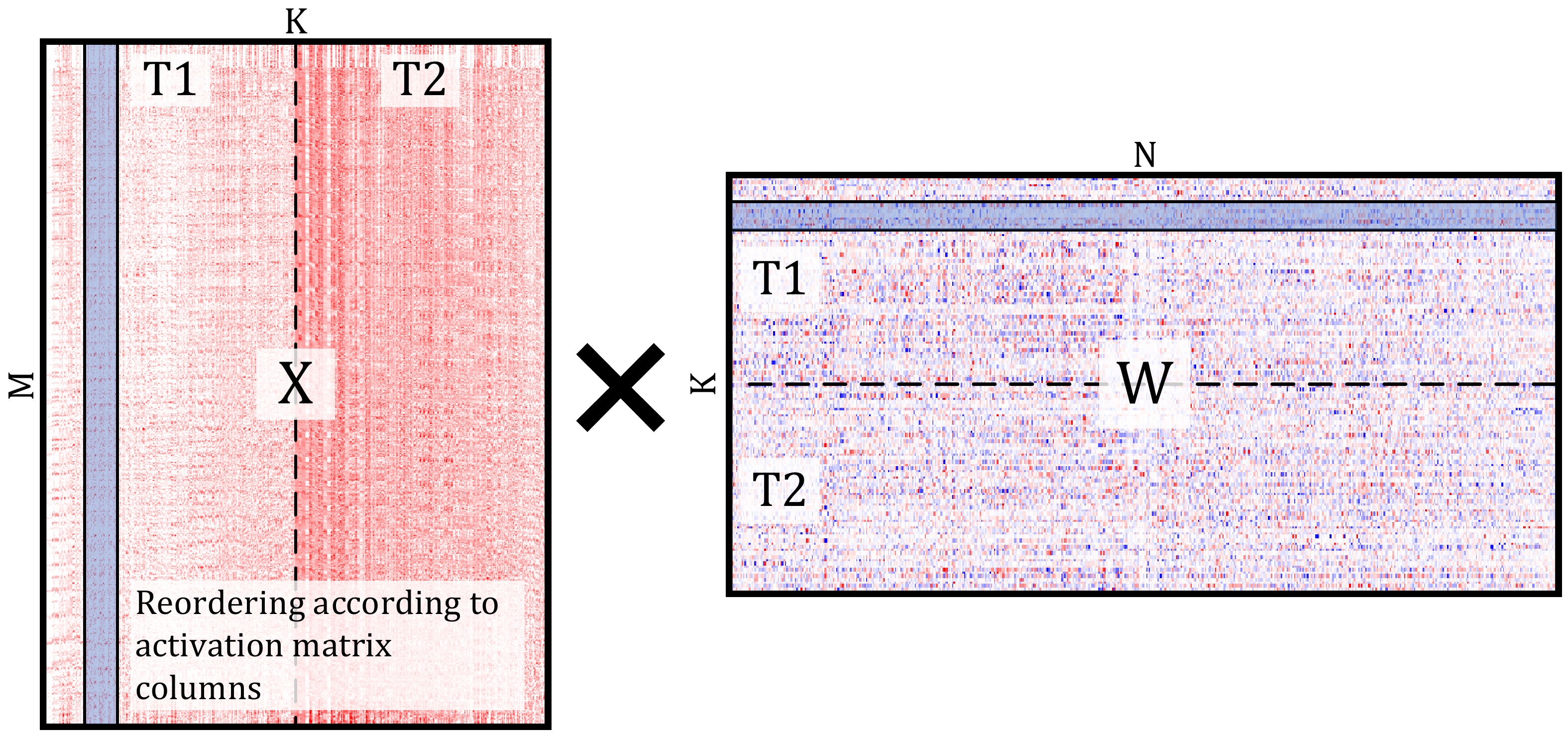}
	  \caption{After reordering}
	  \label{fig:reorder:after}
	\end{subfigure}

\caption{Example of activation and weight tensors from ResNet-18 + ImageNet before and after data arrangement.
         Bold red pixels represent high positive values, bold blue pixels represent high negative values, and white pixels are zeros.
         It is evident that the activation matrix columns are correlated.}
\label{fig:reorder}
\end{figure}

Considering these constraints, we reorder the matrices according to per-layer statistics which are gathered \emph{once} on the activations \cite{zhao2019overwrite, migacz20178}.
Using a random subset of the training set, we log which activation matrix columns are most likely to hold data that is represented by 8-bits.
With these statistics in hand, which at this point are static, the activation matrices are rearranged so that an 8-bit activation value from one thread is more likely to be paired with a zero-valued activation from the other thread, and so that a 4-bit activation value from the one thread is paired with another 4-bit activation from the other thread (Fig.~\ref{fig:reorder}).
In practical terms, during runtime the accelerator will rearrange the layer output according to the pre-determined order for the next layer.
This mechanism is not part of the \ARCH{} core, and is therefore beyond the scope of this paper.

\subsection{PE Microarchitecture}
In addition to the circuitry of a conventional OS-SA PE, the \ARCH{} PE requires additional circuitry:
(1) flexible multiplier units capable of conducting different precision multiplications on demand;
(2) a controller for selecting the right MAC operation as a function of the current input sparsity and/or data width;
(3) on-the-fly precision reduction units; and
(4) additional output buffers, two per thread (for activation and weight).

\subsubsection{Multiplication Decomposition}
\label{sec:fmac}

\begin{figure}[t]
\centering
\includegraphics[width=\columnwidth]{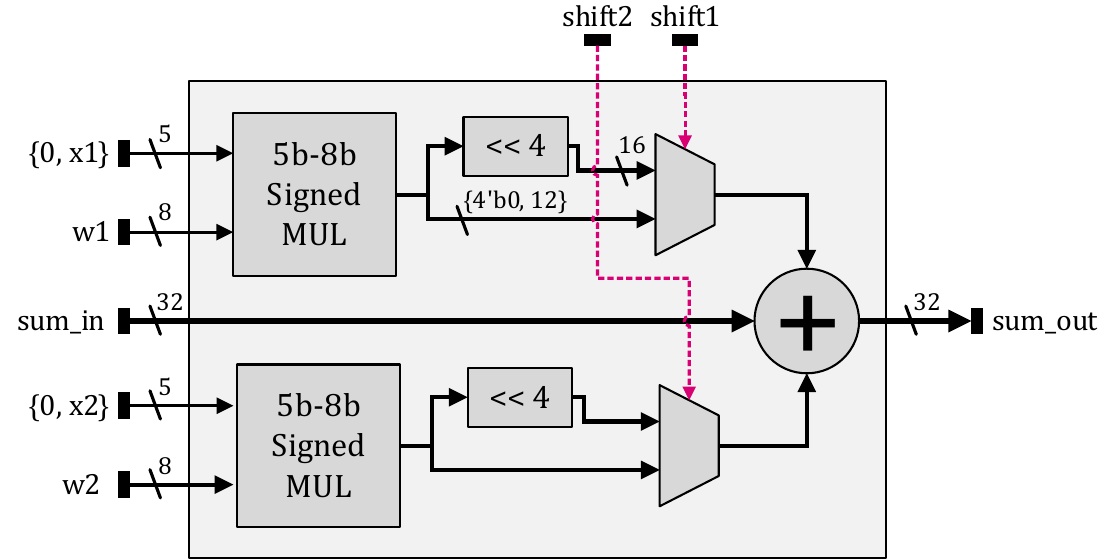}
\caption{Flexible multiplier (fMUL), capable of either one 8b-8b multiplication or two independent 4b-8b multiplications.}
\label{fig:flex_mac_248}
\end{figure}

In the spirit of \cite{sharma2018bit}, we use a flexible multiplier unit (fMUL) capable of multiplication decomposition.
Consider an 8b-8b multiplier and two 8-bit scalar inputs: activations, $\tilde{x}$, and weights, $\tilde{w}$.
We set $\tilde{w}$ to be signed and $\tilde{x}$ to be unsigned, since it is common for activations to follow the ReLU activation function.
Multiplication of these two inputs can be formulated as follows:
\begin{equation}
\begin{aligned}
\tilde{x} \cdot \tilde{w} =& \sum_{i=0}^{7} 2^i x_i \cdot \tilde{w} = \left( \sum_{i=4}^{7} 2^i x_i + \sum_{i=0}^{3} 2^i x_i \right) \cdot \tilde{w} \\
=& \left( 2^4 \sum_{i=0}^{3} 2^{i} x_{i+4} + \sum_{i=0}^{3} 2^i x_i \right) \cdot \tilde{w} \\
=& \,\, (\verb|<<| \; 4)  \underbrace{ ( \{0, \tilde{x}_{\text{MSB}} \} \cdot \tilde{w} )}_{\text{5b-8b sign mult}} +  \underbrace{(  \{0, \tilde{x}_{\text{LSB}} \} \cdot \tilde{w} ) }_{\text{5b-8b sign mult}}  \,,
\end{aligned}
\end{equation}
where we converted $\tilde{x}_{\text{MSB}}$ and $\tilde{x}_{\text{LSB}}$ into two's complement by adding a zero-valued MSB.
This formulation shows that a multiplication of 8-bit unsigned with 8-bit signed can be implemented with two 5b-8b multipliers and a shift.
By adding an additional 4-bit shift to the second 5b-8b signed multiplier and control, we achieve a generalized fMUL that is capable of two independent unsigned-signed 4b-8b multiplications or a single unsigned-signed 8b-8b multiplication, as illustrated in Fig.~\ref{fig:flex_mac_248}.
Moreover, a 4b-8b multiplication can be shifted if the original value was approximated using its 4-bit MSBs.
For example, consider the case illustrated in Fig.~\ref{fig:smt_example:5}.
The multiplier performs a 2x4b-8b multiplication, since the first thread's MSBs are used.
Also, notice that the 4-bit MSBs of the first thread are the input to the multiplier, as opposed to the second thread whose 4-bit LSBs are the input to the multiplier.
The following multiplication then takes place: $1110_2 \cdot 00010111_2 = 322_{10}$ and $0010_2 \cdot 11110010_2 = 484_{10}$.
The first thread computation is followed by a 4-bit shift, which yields a result of $5152_{10} + 484_{10} = 5636_{10}$.

In a similar manner, an 8b-8b multiplication can be formulated as
\begin{equation}
\begin{aligned}
\tilde{x} \cdot \tilde{w} =& \sum_{i=0}^{7} 2^i x_i \cdot \left(   -2^7 w_7 +  \sum_{i=0}^{6} 2^i w_i  \right) \\
=& \left( \sum_{i=4}^{7} 2^i x_i + \sum_{i=0}^{3} 2^i x_i \right) \cdot \\
& \left(   -2^7 w_7 +  \sum_{i=4}^{6} 2^i w_i  + \sum_{i=0}^{3} 2^i w_i \right) \\
=& \,\, (\verb|<<| \; 8)  \overbrace{   ( \{0, \tilde{x}_{\text{MSB}} \} \cdot \tilde{w}_{\text{MSB}} )  }^{\text{5b-4b sign mult}}  +
(\verb|<<| \; 4)   \overbrace{    ( \tilde{x}_{\text{MSB}}  \cdot \tilde{w}_{\text{LSB}} )  }^{\text{4b-4b unsign mult}} + \\
& \,\, (\verb|<<| \; 4)   \underbrace{   ( \{0, \tilde{x}_{\text{LSB}} \} \cdot \tilde{w}_{\text{MSB}} )  }_{\text{5b-4b sign mult}}   + 
\underbrace{ (\tilde{x}_{\text{LSB}} \cdot \tilde{w}_{\text{LSB}} )   }_{\text{4b-4b unsign mult}} \,.
\end{aligned}
\end{equation}
The 8b-8b multiplication can be represented as a combination of two 4b-4b unsigned multiplications and two 5b-4b signed multiplications.
By adding additional shift logic and control, we can generalize this circuit to be capable of performing either four independent 4b-4b multiplications, two independent 4b-8b multiplications, or one 8b-8b multiplication.
The 8b-8b multiplication can be further decomposed or formulated with any other N-bit input variables.
In this paper, however, we only use the two decompositions above, for a 2-threaded and a 4-threaded \ARCH{}.

%
%

\subsubsection{Local PE Control}
\label{sec:pe_ctrl}
The conventional PE array is highly scalable due to lack of control.
Each PE is responsible for a specific task which it conducts locally.
Therefore, to keep \ARCH{} as scalable as conventional SAs are, each PE within \ARCH{} should dynamically control its fMUL locally.
Algorithm~\ref{alg:algorithm1} describes how a 2-threaded PE exploits input sparsity and effective data width (of activations, without loss of generality) to prepare the input data and control for the fMUL unit.
Each cycle, the PE checks the input computation demands versus its available resources, in this case, an 8b-8b fMAC.
If the two threads require the fMAC, the PE checks the data-width of each thread and forwards either its 4-bit MSBs or LSBs.
If one of the threads does not require the fMAC, the PE passes the active thread to the fMUL, enabling an error-free 8b-8b multiplication.
In either case, the PE sets the corresponding shift signals.

\begin{algorithm}[t]
  \caption{--- 2T \ARCH{} PE Logic}\label{alg:algorithm1}
  \begin{algorithmic}[1]
    \Require Two input pairs, $\{x_0, w_0\}$ and $\{x_1, w_1\}$
    \Ensure MAC inputs, $\{\tilde{x}_0, \tilde{w}_0\}$, $\{\tilde{x}_1, \tilde{w}_1\}$, and $\{s_0, s_1\}$ 
    \Statex 
    \For{$\forall~\text{cycle}$}
	    \If{$\text{all arguments are non-zero}$}
	      \For{$\forall~\text{thread}~i$} 
		      \If{$\text{MSBs}(x_i) = \text{4'b0}$} \Comment{4-bit is suffice}
		        \State $\tilde{x}_i \gets \text{LSBs}(x)$
		        \State $s_i \gets 0$
			  \Else     \Comment{On-the-fly quantization}
		        \State $\tilde{x}_i \gets \text{RoundedMSBs}(x_i)$ 
		        \State $s_i \gets 1$ \Comment{Shifted after multiplication}
		      \EndIf
		      
		      \State $\tilde{w}_i \gets w_i$
	      \EndFor
	    \Else
	      \State $i \gets \text{GetActiveThread}(\{x_0, w_0\}, \{x_1, w_1\})$
	      \State $\{\tilde{x}_0, \tilde{w}_0\} \gets \{\text{LSBs}(x_i), w_i\}$
	      \State $\{\tilde{x}_1, \tilde{w}_1\} \gets \{\text{MSBs}(x_i), w_i\}$
	      \State $\{s_0, s_1\} \gets \{0, 1\}$
	    \EndIf
    \EndFor
  \end{algorithmic}
\label{alg:t2_ctrl}
\end{algorithm}

The 4-threaded (4T) implementation considers two more options.
When all four threads require the fMAC, all activations and weights precision are reduced to 4 bits according to the effective input data-width of each thread (i.e., MSBs or LSBs).
For simplicity's sake, a collision of three threads is treated similarly; that is, we reduce all input precision, even though theoretically one thread can utilize a 4b-8b MAC.
Cases of two thread collisions or no collision are handled in the same manner that a 2-threaded \ARCH{} does, but with additional logic that marks the active threads.

\subsubsection{Putting It All Together}
Enabling \VSMT{} does not only require the fMUL (Section~\ref{sec:fmac}) and the control logic (Section~\ref{sec:pe_ctrl}), but also additional activation and weight output registers and on-the-fly precision reduction logic (Fig.~\ref{fig:arch}).
First, the output registers are scaled linearly with the number of threads.
Second, to reduce inputs represented by more than 4 bits to 4 bits, we first round them to the nearest integer that is a whole multiple of 16. 

\ARCH{} data propagates the same as in conventional SA --- each cycle data enters and exits the PEs regardless of the data content.
To maintain such systematic non-blocking operation, we trade backpressure by temporarily reducing numerical precision.
Moreover, avoiding backpressure yields a constant speedup; a 2-threaded and a 4-threaded \ARCH{} will deliver a speedup of 2$\times$ and 4$\times$, respectively.
In the next section, we evaluate the impact of \ARCH{} on accuracy as well as on the hardware.

\section{Evaluation}
\label{sec:eval}

The execution of a DNN layer with two and four threads achieves a constant speedup of 2$\times$ and 4$\times$.
However, the actual impact of \VSMT{} on the underlying hardware, as well as on DNN accuracy, has yet to be evaluated.
In Section~\ref{sec:eval:math}, we describe our evaluation environment and estimate the hardware area, power, and energy of a 16$\times$16 \ARCH{};
in Section~\ref{sec:eval:exp}, we present the results of 2T and 4T \ARCH{}s with five popular CNN models.

\subsection{Methodology}
\label{sec:eval:math}

\textbf{Workloads.}
To evaluate \ARCH{}, we use the ILSVRC-2012 dataset \cite{russakovsky2015imagenet} with five popular CNN models for image classification, as described in Table~\ref{tbl:workloads}.
The models are quantized with a simple 8-bit uniform min-max quantization, using symmetric unsigned quantization for activations and symmetric signed quantization for weights \cite{krishnamoorthi2018quantizing}.
In addition, activations are quantized per layer, whereas weights are quantized per kernel.
This configuration supports an efficient hardware implementation, since each dot product result is multiplied by only two scaling factors: the activations' scaling factor and the corresponding kernel's scaling factor \cite{kravchik2019low}.
Prior to CNN execution, we conduct a quick statistics gathering run on 2K randomly picked images from the training set.
In this rapid phase, we average the min-max values, correct the batch-norm layers' running mean and running variance \cite{sun2019hybrid}, and log the relevant reordering statistics as described in Section~\ref{sec:reorder}; none of these steps involve gradient computation or weight update with gradient descent.
In Section~\ref{sec:eval:exp}, we explore the performance of a 4T \ARCH{} given a pruned network.
For weight pruning, we use simple magnitude-based pruning that iteratively prunes a certain percentage of the model weights followed by retraining, similar to \cite{han2015learning}.

\begin{table}[]

\centering

\begin{tabular}{l|rr|rr}
\hline
\multicolumn{1}{c|}{} & \multicolumn{2}{c|}{\textbf{Accuracy}} & \multicolumn{2}{c}{\textbf{MAC Ops.}} \\
\multicolumn{1}{c|}{} & \multicolumn{1}{c}{\textbf{FP32}} & \multicolumn{1}{c|}{\textbf{INT8}} & \multicolumn{1}{c}{\textbf{CONV}} & \multicolumn{1}{c}{\textbf{FC}} \\ \hline
AlexNet~\cite{krizhevsky2014one} & 56.55\% & 56.36\% & 0.6G & 59M \\
ResNet-18~\cite{he2016deep} & 69.76\% & 69.70\% & 1.8G & 0.5M \\
ResNet-50~\cite{he2016deep} & 76.15\% & 76.24\% & 4.1G & 2M \\
GoogLeNet~\cite{szegedy2015going} & 69.78\% & 69.63\% & 1.5G & 1M \\
DenseNet-121~\cite{huang2017densely} & 74.65\% & 74.66\% & 2.7G & 1M \\ \hline
\end{tabular}%

\caption{The evaluated CNN models (FP32 pre-trained from PyTorch). MAC operations are for a single input image.}
\label{tbl:workloads}
\end{table}

\textbf{CNN simulation.}
We use PyTorch \cite{paszke2019pytorch} to simulate the impact of \ARCH{} on the CNN model accuracies.
Throughout this section, we do not consider the first convolution layer and the fully-connected layers which we leave intact.
The convolution operations are mapped to matrix multiplication operations to fit the hardware simulator \cite{chetlur2014cudnn, kung2018mapping}.




\textbf{Hardware evaluation.}
We implement a 16$\times$16 OS-SA baseline and 16$\times$16 2T and 4T \ARCH{} cores with SystemVerilog.
Synthesis is performed using Synopsys Design Compiler \cite{designcompiler} with the 45nm NanGate open cell \cite{knudsen2008nangate} at a clock frequency of 500MHz (similar to \cite{akhlaghi2018snapea}).
Area and power estimations are extracted from Cadence Innovus \cite{innovus}.
PEs are pipelined internally without affecting the systematic propagation of data between PEs in each cycle.
The pipeline has two stages: the first includes multiplication and control, and the second includes accumulation.
The two-staged pipeline increases latency by a cycle but does not affect throughput.
Table~\ref{tbl:hw_summary} describes the hardware configuration.

\begin{table}[]
         
\centering
\begin{tabular}{lcrr}
\hline
 & \multicolumn{1}{l|}{\textbf{}} & \multicolumn{2}{c}{\textbf{\ARCH{}}} \\
\multicolumn{1}{l|}{} & \multicolumn{1}{c|}{\textbf{SA}} & \multicolumn{1}{c}{\textbf{2T}} & \multicolumn{1}{c}{\textbf{4T}} \\ \hline
\multicolumn{1}{l|}{Array Size} & \multicolumn{3}{c}{16$\times$16 PEs} \\
\multicolumn{1}{l|}{Frequency} & \multicolumn{3}{c}{500MHz} \\
\multicolumn{1}{l|}{Technology} & \multicolumn{3}{c}{45nm \cite{knudsen2008nangate}} \\
\multicolumn{1}{l|}{Throughput [GMACS]} & \multicolumn{1}{r}{256} & $^\dagger$512 & $^\dagger$1024 \\
\multicolumn{1}{l|}{Power [mW] @ 80\% Util.} & \multicolumn{1}{r}{320} & 429 & 723 \\
\multicolumn{1}{l|}{Total Area [mm\textsuperscript{2}]} & \multicolumn{1}{r}{0.220} & 0.317 & 0.545 \\
\multicolumn{1}{l|}{PE [\textmu m\textsuperscript{2}]} & \multicolumn{1}{r}{853} & 1233 & 2122 \\
\multicolumn{1}{l|}{MAC [\textmu m\textsuperscript{2}]} & \multicolumn{1}{r}{591} & 786 & 1102 \\ \hline
\end{tabular}

\caption{Design parameters, power, and area breakdown.
		 PE area includes thread registers, control logic, and the MAC unit.
		 MAC units are two-stage pipelines; their areas include the registers.
         $^\dagger$\ARCH{} throughput is 2$\times$ and 4$\times$ for 2 and 4 threads, respectively, with on-demand precision reduction.}
\label{tbl:hw_summary}
\end{table}

\textbf{Power and energy.}
We estimate power consumption using a synthetic testbench that simulates different SA utilizations.
A utilized PE is defined as a PE with a working MAC unit in any capacity, that is, both operands of at least one input pair are not equal to zero.
To meet a certain utilization operating point, the testbenches zero out the activations at a probability corresponding to a configured utilization.
Activations, rather than weights, are zeroed out, since weights are usually non-zero (when not considering pruning).
The testbenches are used to produce value change dumps (VCDs) that are eventually used to estimate the power of the different SAs.

To estimate energy consumption, we use our PyTorch-based simulator to extract average utilization per layer from each CNN model.
Given the average power of layer $l$, $P_l$, we calculate the energy consumed by layer $l$ as follows:
\begin{equation}
	\begin{aligned}
		E_l = \frac{\text{MAC}_l}{\text{Throughput}} \cdot P_l \,,
	\end{aligned}
\end{equation}
where $\text{MAC}_l$ is the number of MAC operations in layer $l$.
The total model energy is the sum over all layers ($L$): ${E = \sum_{l=1}^{L} E_l}$.
Our evaluation shows that \ARCH{} saves an average of 33\% and 35\% energy when executing the five CNNs with 2 and 4 threads, respectively.

A 2T \ARCH{} does not consume twice the power of a conventional SA.
For example, we estimate that a conventional SA and a 2T \ARCH{} consume 277mW @ 40\% utilization and 429mW @ 80\% utilization, respectively.
Assuming that two threads increase utilization by exactly 2$\times$, the power increase is 1.5$\times$ (429mW @ 80\% divided by 277mW @ 40\%).
Since the 2T \ARCH{} has a constant speedup of 2$\times$ over the conventional SA, energy is reduced.
Energy is further reduced when actual utilization is not doubled or quadrupled, in which case the number of effective MAC operations is reduced.

\subsection{Experimental Results}
\label{sec:eval:exp}

\begin{figure}[t] 
	\centering
	
	\begin{tikzpicture}
	\begin{axis}[
			ybar, bar width=6pt,
			width=8.8cm, height=5.0cm,  
	    	enlarge x limits=0.13, enlarge y limits=0,
	    	xlabel near ticks, ylabel near ticks,
	    	legend style={at={(0.5,-0.20)}, anchor=north,legend columns=-1},
	    	ylabel={Top-1 Accuracy [\%]},
	    	symbolic x coords={AlexNet, ResNet18, ResNet50, GoogLeNet, DenseNet},
	    	xtick=data,
	    	x tick label style = {font=\footnotesize, rotate=0},
			y tick label style = {font=\scriptsize},
			legend style={font=\scriptsize, legend columns=5, legend cell align=left, at={(0.5, 1.01)}, anchor=south, draw=none},
			tick style={draw=none},
			ymajorgrids, major grid style={dashed},
			nodes near coords, ymin={25}, ymax={90},
			every node near coord/.append style={font=\scriptsize, rotate=90, anchor=west},
			label style = {font=\footnotesize},
			set layers=Bowpark
	    ]
	    
	    \addplot[fill=black!60] plot coordinates {(AlexNet,56.4) (ResNet18,69.7) (ResNet50,76.2) (GoogLeNet,69.6) (DenseNet,74.7)};
	    \addplot[fill=black!40] plot coordinates {(AlexNet,53.0) (ResNet18,66.6) (ResNet50,70.1) (GoogLeNet,63.4) (DenseNet,71.9)};
		\addplot[fill=black!30] plot coordinates {(AlexNet,52.3) (ResNet18,50.9) (ResNet50,72.5) (GoogLeNet,41.8) (DenseNet,66.1)};
		\addplot[fill=black!20] plot coordinates {(AlexNet,48.0) (ResNet18,45.3) (ResNet50,63.2) (GoogLeNet,28.9) (DenseNet,60.1)};

		\legend{A8W8 (Baseline), A4W8, A8W4, A4W4}
		
	\end{axis}
	\end{tikzpicture}

\caption{Model robustness to on-the-fly numerical precision reduction of the entire model.
		 The results may be considered as the \emph{worst-case} model performance for two threads (A4W8 and A8W4) and four threads (A4W4).
         Baseline quantization is 8-bit activations and weights (A8W8) with batch-norm recalibration.
         The other quantized operating points derive from the baseline without any calibration, similar to how the PEs perform the precision reduction on-the-fly.}
\label{fig:exp:robustness}
\end{figure}
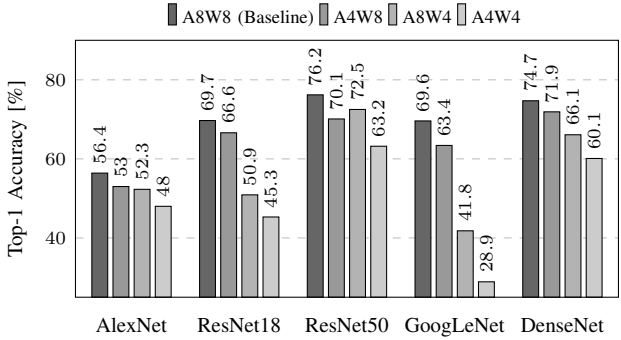

\textbf{Model robustness.}
During inference, \ARCH{} reduces the precision of parts of the computations to 4 bits.
Precision reduction is conducted on-the-fly without variance and bias corrections, which are common in pure quantization mechanisms \cite{finkelstein2019fighting, banner2019post}.
We consider a model as more robust than another if it achieves better accuracy given a decrease in its activations and weights representation \cite{shkolnik2020robust}.
A model whose entire representation was reduced from 8b activations and weights (A8W8) to, for example, 4b activations and 8b weights (A4W8) is equivalent to the worst-case scenario for a 2T \ARCH{}.
It may, therefore, be considered as the lower accuracy boundary for the 2-threaded implementation.
Figure~\ref{fig:exp:robustness} illustrates model robustness to further quantization of the entire model given an A8W8 baseline.
For example, an 8b-8b GoogLeNet whose activations are further quantized to 4 bits incurs a 6.2\% accuracy drop.
We observe that besides ResNet-50, all models are more robust to quantization of their activations rather than their weights.
Therefore, when it is necessary to reduce threads precision, we prefer \ARCH{} to further quantize its activations; we only consider further weight quantization for two-thread collisions with ResNet-50.

\textbf{2T \ARCH{}: sparsity and data-width.}
\VSMT{} exploits 8-bit and 4-bit input sparsity to mitigate noise involved in ``squeezing'' threads into a shared MAC unit.
Table~\ref{tbl:exp:sparse_quant} presents the impact of independently exploiting sparsity (8-bit) and data-width (4-bit) on the different CNN models, given a 2T \ARCH{}.
We denote the different options as follows:
\renewcommand{\labelitemi}{\textendash}
\begin{itemize}
  \item \textbf{S:} exploiting 8-bit sparsity, as illustrated in Fig.~\ref{fig:smt_example:2}.
  \item \textbf{A (W):} exploiting activation (weight) data-width (4-bit) and reducing their precision on-demand, as illustrated in Fig.~\ref{fig:smt_example:3}.
  \item \textbf{Aw (aW):} exploiting activation and weight data-width (4-bit) and reducing activation (weight) precision on-demand, as illustrated in Fig.~\ref{fig:smt_example:4}.
\end{itemize}
As expected, the combination of exploiting sparsity and data-width variability (S+A and S+W) achieves the best results.
Exploiting both activation and weight sparsity (S+Aw and S+aW) does not, however, yield significant or consistent improvement in accuracy.
Therefore, throughout this evaluation we exploit either S+A (for all models) or S+W for ResNet-50.

\begin{table}[t]

\centering
\resizebox{\columnwidth}{!}{%
\begin{tabular}{lrrcrrrr}
\hline
 & \multicolumn{1}{l}{\textbf{}} & \multicolumn{1}{l|}{\textbf{}} & \multicolumn{5}{c}{\textbf{A8W8 \ARCH{}}} \\
\multicolumn{1}{l|}{} & \multicolumn{1}{c}{\textbf{A8W8}} & \multicolumn{1}{c|}{\textbf{min}} & \textbf{S} & \multicolumn{1}{c}{\textbf{A}} & \multicolumn{1}{c}{\textbf{Aw}} & \multicolumn{1}{c}{\textbf{S+A}} & \multicolumn{1}{c}{\textbf{S+Aw}} \\ \hline
\multicolumn{1}{l|}{AlexNet} & 56.36 & \multicolumn{1}{r|}{53.03} & \multicolumn{1}{r}{54.52} & 56.04 & 56.05 & 56.21 & 56.22 \\
\multicolumn{1}{l|}{ResNet-18} & 69.70 & \multicolumn{1}{r|}{66.59} & \multicolumn{1}{r}{67.86} & 67.60 & 67.66 & 68.49 & 68.38 \\
\multicolumn{1}{l|}{GoogLeNet} & 69.63 & \multicolumn{1}{r|}{63.42} & \multicolumn{1}{r}{66.09} & 65.37 & 65.46 & 67.45 & 67.34 \\
\multicolumn{1}{l|}{DenseNet-121} & 74.66 & \multicolumn{1}{r|}{71.94} & \multicolumn{1}{r}{73.45} & 73.00 & 73.23 & 74.05 & 73.87 \\ \hline
 & \multicolumn{1}{c}{} & \multicolumn{1}{c}{} &  & \multicolumn{1}{c}{} & \multicolumn{1}{c}{} & \multicolumn{1}{c}{} & \multicolumn{1}{c}{} \\ \hline
 & \multicolumn{1}{c}{\textbf{}} & \multicolumn{1}{c|}{\textbf{}} & \multicolumn{5}{c}{\textbf{A8W8 \ARCH{}}} \\
\multicolumn{1}{l|}{} & \multicolumn{1}{c}{\textbf{A8W8}} & \multicolumn{1}{c|}{\textbf{min}} & \textbf{S} & \multicolumn{1}{c}{\textbf{W}} & \multicolumn{1}{c}{\textbf{aW}} & \multicolumn{1}{c}{\textbf{S+W}} & \multicolumn{1}{c}{\textbf{S+aW}} \\ \hline
\multicolumn{1}{l|}{ResNet-50} & 76.24 & \multicolumn{1}{r|}{72.49} & \multicolumn{1}{r}{74.36} & 72.36 & 73.00 & 75.10 & 75.22 \\ \hline
\end{tabular}%
}

\caption{Contribution of exploiting sparsity and/or data-width variability to CNNs' top-1 accuracy with a 2T \ARCH{} and without reordering.}
\label{tbl:exp:sparse_quant}

\end{table}

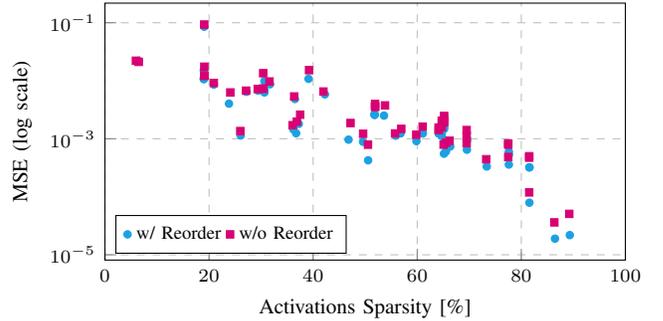
\begin{figure}[t] 
    \centering
	\begin{tikzpicture}
		\begin{axis}[
		title style={font=\footnotesize, at={(0.5, 0.95)}, anchor=south},
		ylabel={MSE (log scale)}, xlabel={Activations Sparsity [\%]}, xmin={0}, xmax={100},
		width=8.5cm, height=5.0cm,
		xtick pos=left, ytick pos=left,
		xlabel near ticks, ylabel near ticks,
		xmajorgrids, ymajorgrids, major grid style={dashed},
		x tick label style = {font=\scriptsize},
		y tick label style = {font=\scriptsize},
		legend style={font=\scriptsize, legend columns=3, at={(0.02, 0.1)}, anchor=west},
		label style = {font=\footnotesize}, ymode = log,
		set layers=Bowpark
		]
		
		\addplot[mark=*, mark size=1.4pt, color=pa1, only marks]  file {plotdata/mse_sparsity_w_reorder.dat};
		\addplot[mark=square*, mark size=1.4pt, color=pa3, only marks]  file {plotdata/mse_sparsity_wo_reorder.dat};
		
		\legend{w/ Reorder, w/o Reorder};
	\end{axis}

	\end{tikzpicture}
	\caption{GoogLeNet MSE due to 2T \ARCH{} as a function of sparsity.
	         Each dot represents a layer.}
	\label{fig:googlenet:mse_sparsity}
\end{figure}

\textbf{2T \ARCH{}: mean squared error (MSE).}
Since GoogLeNet is the one model that exhibits an accuracy drop of more than 1\% (2.18\%), we examine its per-layer MSE due to on-demand precision reduction.
That is, for each layer we compute the MSE between its output, with and without \VSMT{} (without \VSMT{} is considered error-free).
Figure~\ref{fig:googlenet:mse_sparsity} presents the MSE versus the activations sparsity for each GoogLeNet layer, with and without reordering.
MSE and sparsity are correlated, since less sparsity means that more thread collisions occur, and vice versa.

First, we observe that activation reordering decreases the MSE of all layers in GoogLeNet by avoiding thread collisions.
As for classification accuracy, reordering increases the accuracy of ResNet-18, ResNet-50, GoogLeNet, and DenseNet-121 by 0.64\%, 0.24\%, 0.49\%, and 0.35\%, respectively.
Insignificant improvement was recorded in AlexNet.

Second, MSE differs between layers.
\ARCH{} is tunable, and specific layers can be executed with one thread and therefore be error-free.
By doing so, we can trade speedup for accuracy.
To increase GoogLeNet accuracy so as to meet, for example, a 1\% accuracy degradation cap, we execute GoogLeNet without the layer that exhibits the highest MSE.
Since that layer had an insignificant number of MAC operations relative to the rest of the model, we achieve a speedup of 1.98$\times$ with a 69.25\% accuracy --- a 0.38\% degradation from the A8W8 baseline.

\textbf{Accuracy comparison}.
A 2T \ARCH{} enables 8b-8b computations with occasional precision reduction to 4b-8b, which, in a sense, is equivalent to a 4b-8b quantized model.
On the one hand, \ARCH{} is capable of maintaining some of the 8-bit computations, thereby reducing some noise.
On the other hand, precision reduction is conducted on-the-fly, and \ARCH{} lacks the careful quantization parameter adjustments that post-training quantization techniques perform.
We compare 2T \ARCH{} accuracy results to two state-of-the-art post-training quantization methods (Table~\ref{tbl:vdsmt_2t_acc_compare}).
ACIQ \cite{banner2019post} limits the range of the model tensor values by approximating the optimal clipping value analytically from its distribution;
and LBQ \cite{kravchik2019low} proposes an optimization framework to find an optimal quantization parameters of each layer.
\ARCH{} outperforms both ACIQ and LBQ methods in the 4b-8b scenario.
In addition, we expect even less impact on accuracy if weight pruning is considered, as we later demonstrate with a 4T SySMT.


\begin{table}[]

\centering
\resizebox{\columnwidth}{!}{%
\begin{tabular}{l|c|crrc}
\hline
 & \textbf{A/W} & \multicolumn{1}{c}{\textbf{\ARCH{}}} & \multicolumn{1}{c}{\textbf{LBQ}} & \multicolumn{1}{c}{\textbf{ACIQ}} \\ \hline
AlexNet & 4/8 & 56.23 (-0.32) & 55.51 (-1.12)  & 52.30 (-4.32)  \\
ResNet-18 & 4/8 & 69.13 (-0.63) & 68.32 (-1.33) & 68.00 (-1.76)  \\
ResNet-50 & 8/4 & 75.34 (-0.81) & 74.98 (-1.03) & 75.30 (-0.85) \\
DenseNet-121 & 4/8 & 74.40 (-0.25) & 72.31 (-2.16) & \multicolumn{1}{c}{-} \\ \hline
\end{tabular}%
}

\caption{Accuracy comparison of a 2T \ARCH{} (with reordering) versus LBQ~\cite{kravchik2019low} and ACIQ~\cite{banner2019post}.
		 The relative degradation from the relevant FP32 baseline is presented in the round brackets.
		 The input layers are not quantized.
         GoogLeNet was not tested by any of the two comparison methods.}
         
\label{tbl:vdsmt_2t_acc_compare}
\end{table}

\begin{figure}[t] 
    \centering
	\begin{tikzpicture}
		\begin{axis}[
		ylabel={Util. Improvement [$\times$]}, xlabel={Activations Sparsity [\%]},
		width=8.5cm, height=5cm, xmin={0}, xmax={100},
		xtick pos=left, ytick pos=left,
		xlabel near ticks, ylabel near ticks,
		xmajorgrids, ymajorgrids, major grid style={dashed},
		x tick label style = {font=\scriptsize},
		y tick label style = {font=\scriptsize},
		legend style={font=\scriptsize, legend columns=3, at={(0.98, 0.1)}, anchor=east},
		label style = {font=\footnotesize},
		set layers=Bowpark
		]
		
		\addplot[mark=*, mark size=1.4pt, color=pa1, only marks]  file {plotdata/util_sparsity_w_reorder.dat};
		\addplot[mark=square*, mark size=1.4pt, color=pa3, only marks]  file {plotdata/util_sparsity_wo_reorder.dat};
		\addplot[mark=none, dashed, line width=0.5pt][domain=0:100] {0.01*x + 1};
		
		\legend{w/ Reorder, w/o Reorder, Eq.~(\ref{eq:util_gain})};
	\end{axis}

	\end{tikzpicture}
	\caption{GoogLeNet utilization improvement due to \ARCH{} as a function of sparsity, with and without statistical data reordering.
	         Each dot represents a layer.}
	\label{fig:googlenet:util_sparsity}
\end{figure}
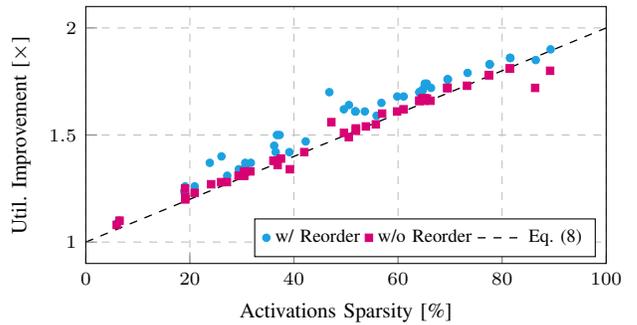

\textbf{2T \ARCH{}: utilization.}
We expect the improvement in utilization and sparsity to be correlated.
Low sparsity means relatively little utilization improvement, since utilization was originally high.
High sparsity means relatively high utilization improvement, since \ARCH{} is capable of ``weaving'' the two threads without collisions.
Figure~\ref{fig:googlenet:util_sparsity} presents the utilization improvement of a 2T \ARCH{} over a conventional SA, with and without reordering, for GoogLeNet as a function of activation sparsity.

The linear trend is somewhat expected.
Consider a single PE, $T$ threads, and inputs $x_i$ and $w_i$ where $i$ is the thread number.
Assume $r_i$ is the probability of $x_i$ to be non-zero and that $w_i$ is always non-zero.
Therefore
\begin{equation}
	\begin{aligned}
		\Pr(\textrm{utilized PE}) 
		&= \Pr(\exists i, x_i w_i \neq 0) \\
		&= 1 - \Pr(\forall i, x_i w_i = 0) \\
		&= 1 - \prod\nolimits_{i=0}^T (1 - \Pr(x_i w_i \neq 0)) \\
		&= 1 - \prod\nolimits_{i=0}^T (1 - r_i) \,.
	\end{aligned}
\end{equation}
For simplicity's sake, we assume all probabilities $r_i$ are equal and that all PEs act the same, so that the PE array utilization is approximately the same as that of a single PE.
Utilization improvement of two threads over one thread may, therefore, be formulated as follows:
\begin{equation}
	\begin{aligned}
		\text{Utilization Gain} = \frac{1 - (1-r)^2}{1 - (1-r)} = s + 1 \,,
	\end{aligned}
\label{eq:util_gain}
\end{equation}
where $s = (1-r)$ is the sparsity.
This result, which is a simple linear curve, is in a good agreement with our measurements (Fig.~\ref{fig:googlenet:util_sparsity}).
Reordering increases utilization of \ARCH{}, since it trades thread collisions with no collisions, thereby increasing utilization while reducing error.
Utilization measurements with reordering are above the analytical line of Eq.~\ref{eq:util_gain}, since the assumption of thread independence does not hold then.

Interestingly, even though utilization is not doubled, a 2T \ARCH{} still achieves 2$\times$ performance.
The mismatch between utilization improvement and performance speedup is due to the PEs' ability to momentarily operate on two low-precision inputs in one cycle by sacrificing some accuracy.


\textbf{2T \ARCH{}: MLPerf.}
MLPerf~\cite{reddi2020mlperf} is becoming a standard as a machine learning benchmark in both academia and industry.
We evaluated \ARCH{} with ResNet-50 and MobileNet-v1~\cite{howard2017mobilenets} checkpoints provided by the MLPerf inference benchmark suite.
ResNet-50 achieves FP32 and 8-bit (with batch-norm running mean and variance recalibration) top-1 accuracy of 76.46\% and 76.42\%, respectively.
To meet ResNet-50 quality target of 99\% defined by MLPerf, we execute two high MSE layers with one thread.
By doing so, a 2T \ARCH{} achieves a speedup of 1.97$\times$ with top-1 accuracy of 75.81\%.
MobileNet-v1 comprises blocks of depthwise convolutions followed by pointwise convolutions.
Since the pointwise convolutions are the bulk of MobileNet-v1 operations, they are executed with two threads, whereas the depthwise convolutions are executed with one thread.
MobileNet-v1 achieves FP32 and 8-bit top-1 accuracy of 71.68\% and 71.41\%, respectively.
Using a 2T \ARCH{}, we achieve a speedup of 1.94$\times$ with top-1 accuracy of 70.68\%, which meets the MLPerf MobileNet-v1 quality target of 98\%.

\textbf{4T \ARCH{}: accuracy comparison.}
As opposed to a 2T \ARCH{}, thread collisions are more likely to occur in a 4T \ARCH{}.
Moreover, thread collision of three or four threads results in the precision reduction of all activations and weights, leading to additional noise.
We therefore examine \ARCH{} operating points, in which some layers execute with two threads instead of four threads, thereby decreasing the noise they contribute during inference.
Layers chosen to be slowed down are those with the highest recorded MSE.
If different layers exhibit approximately the same MSE, we choose to first slowdown those located at the beginning of the network.
Table~\ref{tbl:4t:comparison} presents results for a 4T \ARCH{} and compares them with those of LBQ \cite{kravchik2019low}.
We do not compare our results to those of ACIQ, since ACIQ quantizes its 4-bit activations per-channel, which is not trivial to be implemented in hardware. 
It is worth mentioning that LBQ also considers that some layers exhibit high MSEs and therefore should be treated differently.
For example, LBQ quadruples the number of 4-bit computations in 23.8\% of ResNet-18 layers to achieve their 4-bit results.
We observe that a 4T \ARCH{} achieves competitive results compared with LBQ with only minor algorithmic pre-processing (gathering min-max statistics versus finding an optimal solution to an optimization problem).
Moreover, \ARCH{} is expected to achieve even higher results if pruning is considered.

\begin{table}[]
	     
\centering
\resizebox{\columnwidth}{!}{%
\begin{tabular}{l|rrrr}
\hline
 & \multicolumn{3}{c|}{\textbf{4T \ARCH{}}} & \multicolumn{1}{l}{} \\
 & \multicolumn{1}{c}{\textbf{4T}} & \multicolumn{1}{c}{\textbf{1L@2T}} & \multicolumn{1}{c|}{\textbf{2L@2T}} & \multicolumn{1}{c}{\textbf{LBQ}} \\ \hline
AlexNet & 53.65 (4$\times$) & 56.02 (2.9$\times$) & \multicolumn{1}{c|}{-} & 54.48 \\
ResNet-18 & 64.32 (4$\times$) & 66.08 (3.7$\times$) & \multicolumn{1}{r|}{67.98 (3.5$\times$)} & 67.42 \\
ResNet-50 & 70.79 (4$\times$) & 71.96 (3.9$\times$) & \multicolumn{1}{r|}{72.72 (3.9$\times$)} & 72.60 \\
GoogLeNet & 60.00 (4$\times$) & 64.47 (3.9$\times$) & \multicolumn{1}{r|}{64.83 (3.9$\times$)} & \multicolumn{1}{c}{-} \\
DenseNet-121 & 72.41 (4$\times$) & 72.5 (3.8$\times$) & \multicolumn{1}{r|}{72.82 (3.7$\times$)} & 71.56 \\ \hline
\end{tabular}%
}

\caption{4T \ARCH{} accuracy and speedup with one (1L) and two (2L) layers set to execute at 2T.
	     LBQ also treats layers with high MSE differently.}
\label{tbl:4t:comparison}
\end{table}

\textbf{4T \ARCH{}: weights pruning.}
\ARCH{} exploits unstructured sparsity in both activations and weights.
Yet, conventional DNN training with only a loss function (i.e., no regularization terms) produces weights that do not comprise many zeros.
Different techniques have, therefore, emerged to increase weight sparsity either through regularization, e.g. L1, or by iteratively removing weights and retraining the model to compensate for the degradation in accuracy.
\ARCH{} benefits from sparse inputs, since more zeros means less thread collisions.

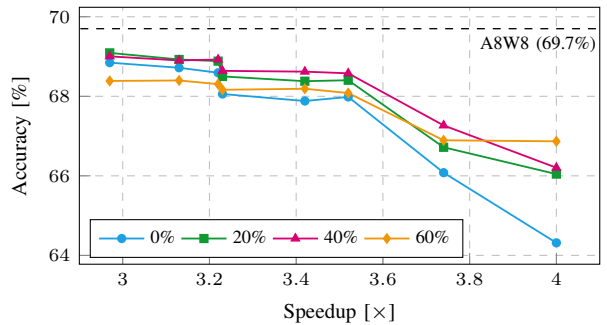
\begin{figure}[t] 
    \centering
	\begin{tikzpicture}
	\begin{axis}[
		ylabel={Accuracy [\%]}, xlabel={Speedup [$\times$]},
		width=8.5cm, height=5cm, xmin={2.9}, xmax={4.1},
		xtick pos=left, ytick pos=left,
		xlabel near ticks, ylabel near ticks,
		xmajorgrids, ymajorgrids, major grid style={dashed},
		x tick label style = {font=\scriptsize},
		y tick label style = {font=\scriptsize},
		legend style={font=\scriptsize, legend columns=4, at={(0.02, 0.1)}, anchor=west},
		label style = {font=\footnotesize},
		set layers=Bowpark
		]
		
		\addplot[mark=*, mark size=1.4pt, color=pa1, line width=0.7pt]  file {plotdata/resnet18_acc_speedup_p0.dat};
		\addplot[mark=square*, mark size=1.4pt, color=pa2, line width=0.7pt]  file {plotdata/resnet18_acc_speedup_p20.dat};
		\addplot[mark=triangle*, mark size=1.6pt, color=pa3, line width=0.7pt]  file {plotdata/resnet18_acc_speedup_p40.dat};
		\addplot[mark=diamond*, mark size=1.6pt, color=pa4, line width=0.7pt]  file {plotdata/resnet18_acc_speedup_p60.dat};
		\addplot[mark=none, dashed, line width=0.5pt][domain=2.5:5] {69.7};
		
		\legend{0\%, 20\%, 40\%, 60\%};
	\end{axis}
	
	\node[below] at (6.1, 3.15) {\scriptsize A8W8 (69.7\%)};

	\end{tikzpicture}
	\caption{ResNet-18 accuracy versus 4T \ARCH{} speedup with different percentage of pruned weights.
	         Each dot represents a measurement with additional layer tuned to run with two threads.}
	\label{fig:4t:acc_speedup}
\end{figure}

Figure~\ref{fig:4t:acc_speedup} presents ResNet-18 accuracy for different percentages of pruned weights (e.g., 20\% means that 20\% of weights are equal to zero).
As before, we trade speedup for accuracy by slowing down layers to run with two threads.   
We observe that with a speedup of 4$\times$, the 60\%-pruned model achieves highest accuracy.
However, as speedup decreases (i.e., as more layers are slowed down), the most pruned model achieves lowest accuracy.
This stems from the fact that the 60\%-pruned model has the lowest baseline accuracy, that is, accuracy of the 60\%-pruned model without \ARCH{} is 68.8\%, 0.9\% below that of original 8-bit model.
Therefore, there is a trade-off between how much accuracy is achieved from avoiding thread collisions thanks to pruning and the potential of the baseline pruned model.

The method in which we set layers from four threads to two threads may not be the most efficient way in terms of accuracy gains to speedup loss.
A different mixture of layers that are set to run with two threads may possibly yield better speedup with better accuracy than those presented in Fig.~\ref{fig:4t:acc_speedup}.
We leave this, however, for future work.

\section{Discussion and Related Work}
\label{sec:related_work}

\textbf{Applying \VSMT{} in other accelerators.}
The concept of \VSMT{} may be beneficial in accelerators other than OS-SAs.
Google's TPU \cite{jouppi2017datacenter, chao2019tpu} is a good candidate, since it is based on an SA core.
However, since TPUv2 comprises FP units in its matrix unit, an NB-SMT solution for FP units is necessary.
With a FP-based core, we expect the relative \VSMT{} overheads to be smaller than those in \ARCH{}.

Eyeriss \cite{chen2016eyeriss} is a 16-bit fixed-point accelerator designated for CNNs.
It exploits the reuse inherent in the convolution operations with a dataflow scheme named row stationary.
Each PE within Eyeriss is capable of executing multiple 2D convolutions in an interleaved fashion.
Instead of interleaving the data, \VSMT{} may be used to execute convolution flows in parallel.
We expect the relative overhead of \VSMT{} to be smaller than that of \ARCH{}, since Eyeriss's PEs are 16-bit, they consist of three pipeline stages, and they have more memory.

DaDianNao \cite{chen2014dadiannao} is largely a pipelined version of the basic computations involved in DNNs, which are multiplications, additions, and activation functions.
The multipliers in the first pipeline stage may be shared and serve several threads, as long as each multiplier does not contribute to different output activations (i.e., psum sharing).
Otherwise, multipliers will need to dynamically propagate their results to the appropriate adder tree as a function of the input values.

SnaPEA \cite{akhlaghi2018snapea} is an architecture with early activation prediction.
Each PE within SnaPEA comprises compute lanes, and each compute lane is responsible for the computation of a single output activation.
Conceptually, an \VSMT{}-enabled SnaPEA may increase compute lane utilization when a thread is predicted to be negative.
In addition, SnaPEA pre-processing phase may also take \VSMT{} into consideration, thereby reducing its impact on accuracy.

\textbf{Exploiting sparsity.}
Sparsity has long been a DNN characteristic exploited in different manners in hardware to achieve performance, power, and energy savings.
(1) Exploiting sparsity with dedicated hardware that operates on compressed encodings, such as SparTen~\cite{gondimalla2019sparten}, Cambricon-S~\cite{zhou2018cambricon}, SCNN~\cite{parashar2017scnn}, EIE~\cite{han2016eie}, and SpArch~\cite{zhang2020sparch};
(2) Predicting whether certain activations in the layer's output are zero-valued and their computation can be, therefore, skipped.
Prediction must be low-cost relative to the original activation computation and can be based on partial computation of the predicted activation \cite{akhlaghi2018snapea, song2018prediction}, on spatially adjacent fully computed activations \cite{shomron2019thanks, shomron2018spatial}, and on input activations \cite{zhu2018sparsenn}, for example;
(3) Skipping over zero-valued input activations and weights \cite{albericio2016cnvlutin}, which can be further enhanced by skipping over zero-valued bits \cite{albericio2017bit, sharify2019laconic};
and (4) Actively fine-tuning the DNN parameters to better fit the hardware architecture \cite{kung2018adaptive, liu2020sta}.
In this paper, we introduce \VSMT{} --- an additional strategy to exploit unstructured sparsity that has yet to be explored.

\textbf{Exploiting data-width variability.}
DNNs can maintain accuracy while altering their numerical precision.
Numerical representation may differ between layers and within layers.
Proteus~\cite{judd2016proteus} and ShapeShifter~\cite{lascorz2019shapeshifter} exploit data-width variability to reduce memory traffic and memory footprint.
Stripes~\cite{judd2016stripes} exploits precision variability with bit-serial computations;
and Bit Fusion~\cite{sharma2018bit} is a configurable accelerator capable of DNN execution in different bit-widths.
In this paper, we demonstrate \VSMT{} with \ARCH{} which exploits data-width variability and DNN tolerance to precision changes in order to ``squeeze'' several threads to the same execution unit in parallel.

\textbf{Multitasking.}
The notion of DNN multitasking has been demonstrated with PREMA~\cite{choi2020prema} and AI-MT~\cite{baek2020multi}.
Both papers decrease latency by prioritizing high-priority tasks, increase resource utilization, and increase memory bandwidth utilization with blocking scheduling algorithms.
\VSMT{}, on the other hand, avoids task blocking by exploiting the resiliency of DNNs, that is, sacrificing computation precision on particular occasions.
In addition, the multitasking demonstrated in this paper with \ARCH{} may be considered as taking place in fine granularities of MAC operations.


\section{Conclusions}
\label{sec:conclusions}
Deep neural networks (DNNs) involve abundant multiply-and-accumulate (MAC) operations, many of which underutilize the underlying hardware due to particular values.
In this paper, we mapped the concept of simultaneous multithreading (SMT), known from CPUs, to hardware designated for DNNs.
We show that by considering the input values and by acknowledging that DNNs may endure some perturbations in their MAC results, non-blocking SMT (\VSMT{}) can increase hardware utilization and save energy with negligible accuracy degradation.
\VSMT{} differs from conventional SMT as it does not stall threads on structural hazards.
Instead, \VSMT{} ``squeezes'' threads to the shared resources when a structural hazard occurs, taking advantage of DNN resiliency and enabling the implementation of multithreading even in rigid structures such as systolic arrays (SAs).
We implement \VSMT{} as an extension to an SA, which we name \ARCH{}, and evaluate its impact on five popular CNN architectures as well as its area, power, and energy.
For example, compared with a conventional SA, a 2-threaded \ARCH{} consumes 1.4$\times$ the area and delivers a 2$\times$ speedup with 33\% energy savings with less than 1\% accuracy degradation.

\section*{Acknowledgment}
We thank the anonymous reviewers for their comments and suggestions.
We also thank Mario Nemirovsky, Yoav Etsion, Shahar Kvatinsky, Ronny Ronen, Tzofnat Greenberg, Moran Shkolnik, Samer Kurzum, and Avi Baum for their valuable feedback.
We acknowledge the support of NVIDIA for its donation of a Titan V GPU used in this research.
This research was supported by Intel-Technion AI center.

\bibliographystyle{IEEEtran}
\bibliography{refs}

\end{document}